\documentclass{article} % For LaTeX2e
\usepackage{collas2023_conference,times}
\usepackage{easyReview}

\usepackage{amsmath,amsfonts,bm}

\def\figref#1{figure~\ref{#1}}
\def\Figref#1{Figure~\ref{#1}}

\def\Secref#1{Section~\ref{#1}}
\def\eqref#1{equation~\ref{#1}}
\def\algref#1{algorithm~\ref{#1}}
\def\Algref#1{Algorithm~\ref{#1}}

\def\1{\bm{1}}

\DeclareMathAlphabet{\mathsfit}{\encodingdefault}{\sfdefault}{m}{sl}
\SetMathAlphabet{\mathsfit}{bold}{\encodingdefault}{\sfdefault}{bx}{n}

\usepackage{hyperref}
\hypersetup{
    colorlinks=true,
    linkcolor=red,
    filecolor=magenta,      
    urlcolor=blue,
    citecolor=purple,
    pdftitle={Overleaf Example},
    pdfpagemode=FullScreen,
    }
\usepackage{todonotes}
\usepackage{graphicx}
\usepackage{url}
\usepackage{notations}
\usepackage{svg}
\usepackage[capitalize,noabbrev]{cleveref}
\usepackage{subcaption}
\usepackage{caption}
\usepackage{color, colortbl}
\usepackage{wrapfig}

\usepackage[linesnumbered,ruled,vlined]{algorithm2e}

\SetCommentSty{mycommfont}
\SetKwInput{KwInput}{Input}                % Set the Input
\SetKwInput{KwOutput}{Output}              % set the Output
\SetKwInput{KwEnv}{Environment}              % set the Output

\newcommand{\algname}{3RL}

\definecolor{Gray}{gray}{0.9}
\definecolor{LightCyan}{rgb}{0.88,1,1}
\usepackage[most]{tcolorbox}
\title{Task-Agnostic Continual Reinforcement Learning: \\Gaining Insights and Overcoming Challenges}

\author{%
  Massimo Caccia$^\dagger$ 
  \thanks{
  corresponding author:   \texttt{massimo.p.caccia@gmail.com} \quad
  $^\dagger$work done while at Amazon Web Services
  }\\
  Mila - Quebec AI Institute \\
  Université de Montréal\\
  \And
  Jonas Mueller$^\dagger$ \\
  Cleanlab\\
  \And
  Taesup Kim$^\dagger$ \\
  Seoul National University\\
  \AND
  Laurent Charlin \\
  Mila - Quebec AI Institute \\
  HEC Montréal\\
  Canada CIFAR AI Chair \\
  \And
  Rasool Fakoor \\
  Amazon Web Services
}

\preprintcopy 

\begin{document}

\maketitle

\begin{abstract}

Continual learning (CL) enables the development of models and agents that learn from a sequence of tasks while addressing the limitations of standard deep learning approaches, such as catastrophic forgetting. In this work, we investigate the factors that contribute to the performance differences between task-agnostic CL and multi-task (MTL) agents. We pose two hypotheses: (1) task-agnostic methods might provide advantages in settings with limited data, computation, or high dimensionality, and (2) faster adaptation may be particularly beneficial in continual learning settings, helping to mitigate the effects of catastrophic forgetting.

To investigate these hypotheses, we introduce a replay-based recurrent reinforcement learning (\algname{}) methodology for task-agnostic CL agents. We assess \algname{} on a synthetic task and the Meta-World benchmark, which includes 50 unique manipulation tasks. Our results demonstrate that \algname{} outperforms baseline methods and can even surpass its multi-task equivalent in challenging settings with high dimensionality. We also show that the recurrent task-agnostic agent consistently outperforms or matches the performance of its transformer-based counterpart.

These findings provide insights into the advantages of task-agnostic CL over task-aware MTL approaches and highlight the potential of task-agnostic methods in resource-constrained, high-dimensional, and multi-task environments.

\end{abstract}

%%%%%%%%%%%%%%%%%%%%%%
\section{Introduction}
%%%%%%%%%%%%%%%%%%%%%%
\label{sec:intro}

Continual learning (CL) creates models and agents that learn from a sequence of tasks. Continual learning agents aim at solving multiple tasks and adapt to new tasks without forgetting the previous one(s), a major limitation of standard deep learning agents~\citep{French99, thrun1995lifelong, mccloskey1989catastrophic, Lesort2019Continual}. 
In many studies, the performance of CL agents is compared against \emph{multi-task} (MTL) agents  trained jointly on all available tasks.
During learning and evaluation, these multi-task agents are typically provided with the identity of the current task (e.g.\ each datum is coupled with its task ID) making them \emph{task-aware}. 
The performance of multi-task agents provides a soft upper bound on the performance of continual learning agents that are bound to learn tasks sequentially and so can suffer from \emph{catastrophic forgetting}~\citep{mccloskey1989catastrophic}. 
Moreover, continual learning agents are often trained without task IDs, a challenging setting motivated by practical constraints and known as \emph{task-agnostic} CL~\citep{zeno2019task,He2019TaskAC,caccia2020online,Berseth2021CoMPSCM}. 

In this work, our aim is to understand the factors that explain the difference in performance between continual-learning methods and  their multi-task counterparts. 
We hypothesize that both task-agnosticity and continual learning may provide advantages when learning from limited data and computation or in settings where the dimensionality of (the observation space of) each task and the number of tasks to solve are high.
Our reasoning is as follows. 
Task-agnostic methods may learn the ability to adapt more quickly to novel environments that are similar to previous environments. 
In comparison, task-aware methods may rely on task ID and so learn to memorize environments. As a result, they may require more data or computations to adapt to other environments (Hypothesis 1).
We further hypothesize that this faster adaptation may be particularly beneficial in continual-learning settings, where it can help combat the impact of catastrophic forgetting (Hypothesis 2).

\begin{figure}[t]
  \centering
    \includegraphics[width=0.95\linewidth]{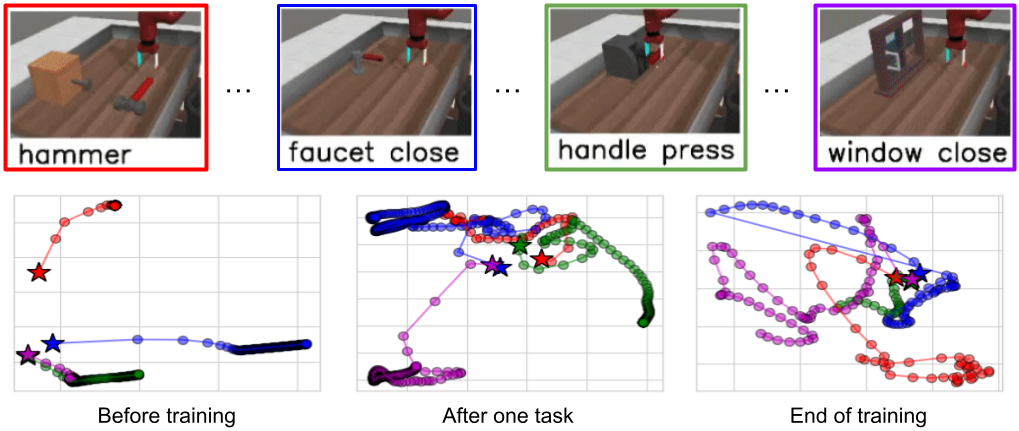} 
    \caption[]
    {\small{\textbf{The Continual-World\footnotemark benchmark (top) as well as evolving RNN representations (bottom).} Continual World consists of 10 robotic manipulation environments (four are shown above) within the same state space and built on a common reward structure composed of shared components: reaching, grasping, placing and pushing. 
    We explore the task-agnostic setting in which agents need at least part of a trajectory (rather than a single state) for task identification. 
    We performed a PCA analysis of 3RL's RNN representations at different stages of training. 
    One episode is shown per task, and the initial task representation (at $t=0$) is indicated by a star. 
    As training progresses, the model learns i) a task-invariant initialization, drawing the initial states closer together, and ii) richer, more diverse representations. 
    Furthermore, the representations constantly evolve throughout the episodes, suggesting the RNN performs more than just task inference: it provides useful local information to the policy and critic.
    }}
    \label{fig:fig1}
\end{figure}

To evaluate the above hypotheses, we instantiate several task-aware and task-agnostic methods.
For the task-agnostic continual-learning agents, we add a recurrent memory and the capability of replaying the trajectory of previous tasks \citep{rolnick2019experience}.
We refer to this methodology as \emph{replay-based recurrent reinforcement learning} (\algname{}) (see the bottom of \Cref{fig:fig1} for a visualization of its learned representations).

We evaluate these methods on two benchmarks. The first consists of a synthetic task in which agents learn to maximize a quadratic function, with controllable dimensions for the observation and action spaces. The empirical results on this benchmark provide validation for Hypothesis 1. They show that \algname{} outperforms baseline methods, particularly in terms of robustness, as we reduce data and compute resources or increase the dimensionality of the problem. The results also support Hypothesis 2 and even demonstrate that \algname{} can outperform its multi-task equivalent, especially in challenging settings with high dimensionality.
We explain how the observed improvement could be attributed to a decrease in gradient conflict \cite{yu2020gradient}.% 
Furthermore, we compare \algname{} with its transformer-based equivalent \citep{vaswani2017attention}, and find that the recurrent task-agnostic agent either outperforms or equals the performance of the transformer-based approach.

Subsequently, we assess our method's performance on the Meta-World benchmark~\citep{yu2019meta}, which comprises 50 unique manipulation tasks. Our findings further validate both hypothesis and demonstrate that \algname{} outperforms other CRL methods and successfully attains its MTL soft-upper bound—an encouraging accomplishment that, to the best of our knowledge, no other CRL method has achieved thus far.\footnote{It should be noted that PackNet has been reported to surpass multi-task baselines. In this context, we assert that \algname{} can outperform its multi-task counterpart. PackNet, however, lacks an MTL equivalent for comparison purposes.}
The code to reproduce the results is publicly available.\footnote{\url{https://github.com/amazon-science/replay-based-recurrent-rl}}

%% figure at https://docs.google.com/drawings/d/1b6xix4MDYgEtyLFnIm0M2-rEtZvIBNqy_IKq6zCaRzk/edit
%% and https://docs.google.com/drawings/d/1F9wzXk8catBoihkVuwbCAoCW7YvjUBAwHqBKoD0sKX8/edit

\footnotetext{The figures depict the rendering of Meta-World, and not what the agent observes. The agent's observation space is composed of objects, targets, and gripper positions. Because of the randomness of those positions, the agents need more than one observation to properly infer the hidden state.}

%%%%%%%%%%%%%%%%%%%%%%%%%%%%%%%%%%%%%%%%%%%%%%%%%%%%%%%%%
\section{Background \& Task-agnostic Continual Reinforcement Learning (TACRL)}
%%%%%%%%%%%%%%%%%%%%%%%%%%%%%%%%%%%%%%%%%%%%%%%%%%%%%%%%%
\label{sec:tacrl}

Here, we define TACRL, and contrast it against multi-task RL as well as task-aware settings. 

%----------------------------------
\paragraph{MDP.} 
The RL problem is often formulated using a Markov decision process (MDP)~\citep{PutermanMDP1994}. 
An MDP is defined by the five-tuple $\langle\ST, \AS, \TP, r, \gamma \rangle$ with $\ST$ the state space, $\AS$ the action space, $\TP(s'| s,a)$ the transition probabilities, $r(s,a) \in \mathbb{R} $ or equivalently $r_i$ the reward obtained by taking action $a \in \AS$ in state $s \in \ST$, and $\gamma \in [0,1)$ a scalar constant that discounts future rewards.
In RL, the transition probabilities and the rewards are typically unknown and the objective is to learn a policy, $\pi(a|s)$ that maximizes the sum of discounted rewards $\mathcal{R}^\pi_t = \sum_{i=t}^\infty \gamma^{i-t} r_i$ generated by taking a series of actions $a_t \sim \pi(\cdot|s_t)$.
The Q-value $Q^\pi(s,a)$ corresponding to a policy $\pi$, is defined as the expected return starting at state $s$, taking $a$, and acting according to $\pi$ thereafter: $ Q^\pi(s,a) = \underset{\pi}{\EE} \Big[\sum_{t=0}^\infty \gamma^{t} r_t \Big] =  r(s, a) + \gamma \underset{s', a'}{\EE}\Big[ Q^\pi(s', a') \Big]$.
\iffalse
\begin{align*}
    Q^\pi(s,a) & = \underset{\pi}{\EE} \Big[\sum_{t=0}^\infty \gamma^{t} r_t \Big] =  r(s, a) + \gamma \underset{s', a'}{\EE}\Big[ Q^\pi(s', a') \Big]
    %\label{eq:realq}
\end{align*}
\fi

%----------------------------------
\paragraph{POMDP.}
% \TodoOld{sprinkle more POMDP motivation. Can use https://arxiv.org/pdf/2006.10701.pdf for inspiration}
In most real-world applications, if not all, the full information about an environment or a task is not always available to the agent due to various factors such as limited and/or noisy sensors, different states with identical observations, occluded objects,  etc.~\citep{LITTMAN1995362,fakoorpomdp}. 
For this class of problems in which environment states are not fully observable by the agent, partially-observable Markov decision processes (POMDPs)~\cite{kaelbling1998planning} are used to model the problem.
A POMDP is defined by a seven-tuple   $\langle\ST, \AS, \TP, \OS, \OP, r, \gamma \rangle$ that can be interpreted as an MDP augmented with an observation space $\OS$ and an observation-emission function $\OP(x'|s)$. 
In a POMDP, an agent cannot directly infer the current state of the environment $s_t$ from the current observation $x_t$. 
We split the state space into two distinct parts: the one that is observable $x_t$, which we refer to as $s^o_t$, and the remainder as the hidden state $s^h_t$, similarly to~\citep{Tianwei2021recurrent}.
To infer the correct hidden state, the agent has to take its history into account: the policy thus becomes  $\pi(a_t | s^o_{1:t}, a_{1:t-1}, r_{1:t-1} )$.
An obvious choice to parameterize such a policy is with a recurrent neural network~\citep{LongJi1993,WHITEHEAD1995271, Bakker2001,fakoor2019meta,Tianwei2021recurrent}, as described in \Cref{sec:method}.
The objective in partially observable environments is also to learn a policy that maximizes the expected return $\underset{s^h}{\EE} \Big[ \underset{\pi}{\EE} \big[\sum_{t=0}^\infty \gamma^{t} r_t \big] | s^h \Big]$.

%---------------------------------------------------------
\paragraph{Task-agnostic Continual Reinforcement Learning (TACRL).} 
TACRL agents operate in a POMDP special case, explained next, designed to study the \emph{catastrophic forgetting} that plagues neural networks~\citep{mccloskey1989catastrophic} when they learn on non-stationary data distributions, as well as \emph{forward transfer} \citep{ContinualWorld,bornschein2022nevis,caccia2023towards}, i.e., a method's ability to leverage previously acquired knowledge to improve the learning of new tasks \citep{lopez2017gradient}. 
First, TACRL's environments assume that the agent does not have a causal effect on $s^h$.
This assumption increases the tractability of the problem. 
It is referred to as a hidden-mode MDP (HM-MDP)~\citep{choi2000hidden}. \tblref{tab:tacrl} provides and compares its mathematical description to other settings.

%The following assumptions help narrow down the forgetting problem and knowledge accumulation abilities of neural networks.
TACRL's assumes that $s^h$ follows a non-backtracking chain.  
Specifically, once the chain reaches a hidden state it stays in that state for a fixed number of steps. Further previously visited states are never revisited.
The canonical evaluation of TACRL evaluates the  \emph{global return}, that is the performance of the approaches across all tasks (refer to Table \ref{tab:tacrl} under the evaluation column of TACRL for the specific formula). 
This is different than the \emph{current return} metric that only reflects performance on the immediate task at hand. Using the global return,
we can tell precisely which algorithm has accumulated the most knowledge about all hidden states at the end of its training.
In TACRL, the hidden states can be encoded into a single categorical variable often referred to as \emph{context}, but more importantly in CRL literature, it represents a \emph{task}.
As each context can be reformulated as a specific MDP, we treat tasks and MDPs as interchangeable.

\paragraph{Task Awareness}
In practical scenarios,  deployed agents  cannot always assume full observability, i.e.\ access to a \emph{task label} or \emph{ID} indicating exactly which task they are solving or analogously which state they are in.
They might not even have the luxury of being ``told'' when the task changes (unobserved task boundary) in which case agents might have to infer it from data. This setting is known as \emph{task agnosticism} \citep{zeno2019task}. 
Although impractical, CL research often uses \emph{task-aware} methods, which observe the task label, as soft upper bounds for task-agnostic methods~\citep{van2019three,zeno2019task}.
Augmented with task labels, the POMDP becomes fully observable, in other words, it is an MDP.

\paragraph{Multi-task Learning (MTL).}
% CL is plagued by catastrophic forgetting, in which agents quickly forget previously acquired knowledge about old tasks when learning a new task. 
For neural network agents, catastrophic forgetting results from the stationary-data-distribution assumption of stochastic gradient descent being violated. As a result, the network parameters become specific to data from the most recent task.
Thus it is generally preferable to train on data from all tasks jointly as in MTL \citep{mtsurvey}. 
However this may not be possible in many settings, and CL is viewed as a more broadly applicable methodology that is expected to perform worse than MTL \citep{rolnick2019experience, Chaudhry18}.

For RL specifically, multi-task RL (MTRL) often refers to scenarios with families of similar tasks (i.e.\ MDPs) where the goal is to learn a policy (which can be contextualized on each task's ID) that maximizes returns across \emph{all} the tasks~\citep{yang2020multi, calandriello2014multitask, Kirkpatrick3521}. 
While seemingly similar to CRL, the key  difference is that MTRL assumes data from all tasks are readily available during training and each task can be visited as often as needed. 
These are often impractical requirements, which CRL methods are not limited by. 
\Cref{tab:tacrl} summarizes the settings we have discussed in this section. Refer to \appref{app:pomdp_further} for further details and related settings.

\begin{center}
\tabcolsep=2pt\relax
 \scriptsize%
 \makebox[\textwidth]{\begin{tabular}{cccccc}
        & T  & $\pi$ & Objective & Evaluation \\
        \hline \hline \\
        MDP \citep{Sutton1998} & $p(s_{t+1} | s_t, a_t)$ & $\pi(a_t | s_t)$ & $\underset{\pi}{\EE} \big[\sum_{t=0}^\infty \gamma^{t} r_t \big]$ & same as Objective \\
        \rowcolor{LightCyan}
        POMDP \citep{kaelbling1998planning} & $p(\blue{s_{t+1}^h, s_{t+1}^o} | \blue{s_t^h, s_t^o}, a_t)$ & $\pi(a_t | \blue{s^o_{1:t}, a_{1:t-1}, r_{1:t-1}})$ & $\underset{\blue{s^h}}{\EE} \Big[ \underset{\pi}{\EE} \big[\sum_{t=0}^\infty \gamma^{t} r_t \big] | \blue{s^h} \Big]$ & same as Objective \\
        HM-MDP \citep{choi2000hidden} & $ p(s_{t+1}^o | \blue{s_{t+1}^h}, s_t^o, a_t) p(\blue{s_{t+1}^h}
        | s_t^h)$ & $\pi(a_t | s^o_{1:t}, a_{1:t-1}, r_{1:t-1})$ & $\underset{s^h}{\EE} \Big[ \underset{\pi}{\EE} \big[\sum_{t=0}^\infty \gamma^{t} r_t \big] | s^h \Big]$ & same as Objective \\
        \rowcolor{LightCyan}
        \textbf{Task-agnostic CRL} & $ p(s_{t+1}^o | s_{t+1}^h, s_t^o, a_t) p(s_{t+1}^h | s_t^h)$ & $\pi(a_t | s^o_{1:t}, a_{1:t-1}, r_{1:t-1})$ & $\underset{s^h}{\EE} \Big[ \underset{\pi}{\EE} \big[\sum_{t=0}^\infty \gamma^{t} r_t \big] | s^h \Big]$ & $\underset{\blue{\tilde{s}^h}}{\EE} \Big[ \underset{\pi}{\EE} \big[\sum_{t=0}^\infty \gamma^{t} r_t \big] | s^h \Big]$ \\
        Task-Aware CRL & $ p(s_{t+1}^o | s_{t+1}^h, s_t^o, a_t) p(s_{t+1}^h | s_t^h)$ & $\pi(a_t | \blue{s^h_t, s^o_t})$ & $\underset{s^h}{\EE} \Big[ \underset{\pi}{\EE} \big[\sum_{t=0}^\infty \gamma^{t} r_t \big] | s^h \Big]$ & $\underset{\tilde{s}^h}{\EE} \Big[ \underset{\pi}{\EE} \big[\sum_{t=0}^\infty \gamma^{t} r_t \big] | s^h \Big]$ \\
        \rowcolor{LightCyan}
        Multi-task RL & $ p(s_{t+1}^o | s_{t+1}^h, s_t^o, a_t) p(\blue{s_{t+1}^h})$ & $\pi(a_t | s^h_t, s^o_t)$ & $\underset{\blue{\tilde{s}^h}}{\EE} \Big[ \underset{\pi}{\EE} \big[\sum_{t=0}^\infty \gamma^{t} r_t \big] | s^h \Big]$ & same as Objective  \\
        % Meta RL & $ p(s_{t+1}^o | s_{t+1}^h, s_t^o, a_t) p(s_{t+1}^h)$ & $\pi(a_t | s^o_{1:t}, a_{1:t-1}, r_{1:t-1})$ & $\underset{\blue{\tilde{s}^h_{\text{train}}}}{\EE} \Big[ \underset{\pi}{\EE} \big[\sum_{t=0}^\infty \gamma^{t} r_t \big] | s^h_{\text{train}} \Big]$ & $\underset{\blue{\tilde{s}^h_{\text{test}}}}{\EE} \Big[ \underset{\pi}{\EE} \big[\sum_{t=0}^\infty \gamma^{t} r_t \big] | s^h_{\text{test}} \Big]$ \\
        % Continual Meta-RL & $ p(s_{t+1}^o | \blue{s_{t+1}^h}, s_t^o, a_t) p(\blue{s_{t+1}^h}
        % | s_t^h)$ & $\pi(a_t | s^o_{1:t}, a_{1:t-1}, r_{1:t-1})$ & $\underset{s^h_{\text{train}}}{\EE} \Big[ \underset{\pi}{\EE} \big[\sum_{t=0}^\infty \gamma^{t} r_t \big] | s^h_\text{train} \Big]$ & $\underset{\blue{\tilde{s}^h_{\text{test}}}}{\EE} \Big[ \underset{\pi}{\EE} \big[\sum_{t=0}^\infty \gamma^{t} r_t \big] | s^h_{\text{test}} \Big]$ \\
    \end{tabular}}
    \captionof{table}{\small{Summarizing table of the settings relevant to TACRL. 
    For readability purposes, $\tilde{s}^h$ denotes the joint distribution of all $s^h$. 
    %The Evaluation column is left blank when it equals to the Objective column.
    The blue colorization highlights the changes from the previous setting to the next.
POMDP serves as the most general framework, with MDP being a special case where no hidden states exist. HM-MDP is also a special case of POMDP, characterized by non-stationary hidden states. We then introduce TACRL as a specific instance of HM-MDP, featuring a non-backtracking chain for hidden states and employing the global return on all hidden states as the evaluation metric. Task-Aware CRL closely resembles TACRL but with full observability, while MTRL is akin to Task-Aware CRL, but with stationary hidden states.
}}
    \label{tab:tacrl}
\end{center}

%%%%%%%%%%%%%%%%%%%%%%%%%
\section{Methods \& Hypotheses}
%%%%%%%%%%%%%%%%%%%%%%%%%
\label{sec:method}

In this section, we detail the base algorithm and different model architectures used for assembling different continual and multi-task learning baselines. 
We also put forth some hypotheses that pertain to the aptitude of diverse modeling approaches to exhibit superior or inferior performance across varied learning scenarios.

%%%%%%%%%%%%%%%%%%%%%%%
\subsection{Algorithms}
\label{ssec:algo}

We use off-policy RL approaches which have two advantages for (task-agnostic) continual learning. 
First, they are more sample efficient than online-policy ones~\citep{haarnoja2018soft, fakoor2019p3o}.
Learning from lower-data regimes is preferable for CRL since it is typical for agents to only spend short amounts of time in each task and for tasks to only be seen once.
Second, task-agnostic CRL most likely requires some sort of replay function \citep{Traore19DisCoRL,Lesort2019RegularizationSF}. 
This is in contrast to task-aware methods which can, at the expense of computational efficiency, \emph{freeze-and-grow}, e.g.\ PackNet \citep{mallya2018packnet}, incur no forgetting.
Off-policy methods, by decoupling the learning policy from the acting policy, support the replaying of past data. 
In short, off-policy learning is the approach of choice in CRL.\footnote{Note that our findings are not limited to off-policy methods, in fact, our \algname{} model can be extended to any on-policy method as long as it utilizes a replay buffer~\citep{fakoor2019p3o}. Having the capability to support a replay buffer is more important than being on-policy or off-policy.}

%------------------------
\paragraph{Base algorithm.}
%The Soft Actor-Critic (SAC)~\citep{sachaarnoja18b} is an off-policy actor-critic algorithm for continuous actions. SAC adopts a maximum entropy framework that learns a stochastic policy which maximizes the expected return and also encourages the policy to contain some randomness. To accomplish this, SAC utilizes an actor/policy network $\pi_\phi$ and critic/Q network $Q_\theta$, parameterized by $\phi$ and $\theta$ respectively. Q-values are learnt by minimizing one-step temporal difference (TD) error by sampling previously collected data from the replay buffer~\citep{lin1992self}.
%For more details on SAC, please look at \appref{app:sac}.

We utilize Soft Actor-Critic (SAC)~\citep{sachaarnoja18b} as the base algorithm in this paper. SAC is an off-policy actor-critic method for continuous control, where it learns a stochastic policy that maximizes the expected return while also encouraging the policy to contain some randomness. It uses an actor/policy network $\pi_\phi$ and critic/Q network $Q_\theta$, which are parameterized by $\phi$ and $\theta$, respectively. Refer to \appref{app:sac} for more details.

%%%%%%%%%%%%%%%%%%%%%%%%%
\subsection{Models}
\label{ssec:algo}

%We consider various architectures to handle multi-task learning (MTL) as well as continual learning (CL) in both task-aware and task-agnostic setting.

We explore different architectures for MTL and CL in both task-aware and task-agnostic settings.

\paragraph{Task ID modeling (TaskID).} We assume that a model such as SAC can become task adaptive by providing task information to the networks. 
Task information such as task ID (e.g.\ one-hot representation), can be fed into the critics and actor networks as additional input:
$Q_\theta(s, a, \tau)$ and $\pi_\phi(a|s,\tau)$ where $\tau$ is the task ID. 
We refer to this baseline as Task ID modeling (TaskID).
This method is applicable in both multi-task learning and continual learning.
% Note that to properly train the model with this setting, the maximum number of tasks to be learned has to be defined a priori.

\paragraph{Multi-head modeling (MH).} For multi-task learning (which is always task-aware), the standard SAC is typically extended to have multiple heads \citep{yang2020multi,yu2019meta, wolczyk2021continual, yu2020gradient}, where each head is responsible for a single distinctive task, i.e. $Q_\Theta = \{Q_{\theta_k}\}_k^{K}$ and $\pi_\Phi = \{\pi_{\phi_k}\}_k^{K}$ with $K$ the total number of tasks. 
MH is also applicable to all reinforcement learning algorithms. That way, the networks can be split into 2 parts: (1) a shared state representation network (feature extractor) and (2) multiple prediction networks (heads). 
This architecture can also be used for task-aware CL, where a new head is newly attached (initialized) when an unseen task is encountered during learning.

We also employ this architecture in the task-agnostic setting for both MTL and CL. Specifically, we fix the number of heads a priori to the total number of tasks and select the most confident actor head based on the policy's entropy and the most optimistic critic head. Task-agnostic multi-head (TAMH) enables us to quantify the potential MH gains over the base algorithm: the improvements of MH over TAMH are strictly attributed to the advantage of task information and not due to the increased capacity resulting from having multiple heads.

\paragraph{Task-agnostic recurrent modeling (RNN).} Recurrent neural networks are capable of encoding the history of past data~\citep{LongJi1993,WHITEHEAD1995271, Bakker2001,fakoor2019meta, Tianwei2021recurrent}. 
Their encoding can implicitly identify different tasks (or MDPs). 
Thus, we introduce RNNs as a history encoder where the history is defined by $\{(s_i,a_i,r_i)\}_i^N$ and we utilize the last hidden state $z$ as additional input data for the actor~$\pi_\phi(a|s,z)$ and critic~$Q_\theta(s, a, z)$ networks. 
This allows us to train models without any explicit task information, and therefore we use this modeling, especially for task-agnostic continual learning. More RNN details are provided later in the next subsection.

There is a fundamental distinction between two primary approaches: task-aware and task-agnostic. 
Although both aim to address the challenge of solving multiple tasks, they differ in their strategies for achieving this objective.
Task-aware methods may rely on task identity to solve tasks and as a result, memorize a solution specific to each task.
Essentially, they attempt to learn and remember the optimal policy for all tasks by sharing parameters in a single neural network. 
Task-agnostic approaches, on the other hand, aim to achieve the same goal without complete memorization.

Instead, task-agnostic methods may focus on learning a general solution that can perform \emph{fast adaptation} \citep{finn2017model} to previous and new tasks via a task-inference module—typically an RNN—as demonstrated in \cite{Tianwei2021recurrent}.
We can illustrate the distinction using the example of a general-purpose robot. A task-aware robot would learn and remember the best manipulation approach for each individual object, while a task-agnostic robot would instead learn a general manipulation policy that can quickly adjust to each object's characteristics based on proprioceptive and other feedback.
However, the optimal approach may vary depending on the specific scenario, such as the number of objects and their variability in terms of physical properties. This line of reasoning leads us to formulate a first hypothesis.

\begin{tcolorbox}[enhanced,attach boxed title to top center={yshift=-3mm,yshifttext=-1mm},
  colback=blue!5!white,colframe=blue!75!black,colbacktitle=red!80!black,
  title=Hypothesis 1,fonttitle=\bfseries,
  boxed title style={size=small,colframe=red!50!black} ]
    When the reward and transition function share a structure across multiple tasks, task-agnostic approaches have the potential to outperform task-aware approaches in settings where task memorization is difficult, e.g., when the dimensionality and number of tasks increase, or when data and compute resources are limited.
\end{tcolorbox}

\paragraph{Task-agnostic attention-based modeling (TX).} Transformers~\citep{vaswani2017attention} have demonstrated remarkable success in a wide range of domains, including natural language processing and computer vision. These models employ self-attention mechanisms to process input sequences, enabling them to capture long-range dependencies more effectively than RNNs. We use Transformer-based architectures as an alternative history encoder for TACRL. In this setup, the history definition remains consistent with RNNs, and the Transformer-encoded representation of the last timestep is similarly fed to the actor and critic networks. Fundamentally, TX and RNN should approximate a similar solution, but comparing their performance can inform us as to their relative performance.
% by $\{(s_i,a_i,r_i)\}_i^N$, and we leverage the output representations as additional input data for the actor~$\pi_\phi(a|s,z)$ and critic~$Q_\theta(s, a, z)$ networks, where $z$ denotes the Transformer-encoded representation of the last token. 
% This approach allows us to exploit the advantages of attention-based models, such as efficient parallelization and improved long-range dependency modeling, while maintaining the task-agnostic property. 

%%%%%%%%%%%%%%%%%%%%%%%
\subsection{Baselines}
\label{ssec:approaches}
\textbf{FineTuning} is a simple approach to a CL problem. It learns each incoming task without any mechanism to prevent forgetting. 
Its performance on past tasks indicates how much forgetting is incurred in a specific CL scenario.

\textbf{Experience Replay (ER)} accumulates data from previous tasks in a buffer for retraining purposes, thus slowing down forgetting by simulating a multi-task setting \citep{rolnick2019experience,Aljundi2019Online,Chaudhry2019ContinualLW,lesort2020continual}. 
Although simple, it is often a worthy adversary to recent and complex CL methods.
One limitation of replay is that, to approximate the data distribution of all tasks, its compute requirements scale linearly with the number of tasks, leaving little compute for solving the current task, assuming a fixed compute budget.
To achieve a better trade-off between remembering previous tasks and learning the current one, we use a strategy that caps replay by oversampling the data captured in the current task, as explained in \Algref{alg:3rl} L8-9. Note that to do this we maintain two separate buffers, which requires, a priori, being task-aware. However, the desired behavior can be achieved in a task-agnostic setting by oversampling recently collected data.

\textbf{Multi-task (MTL)} trains on all tasks simultaneously and so it does not suffer from the challenges arising from learning on a non-stationary task distribution. It serves as a soft upper bound for CL methods.

\textbf{Independent} learns a separate model for each task, effectively eliminating the challenges associated with CL and MTL, such as learning with conflicting gradients~\citep{yu2020gradient}. However, due to its design, it will never achieve transfer in terms of an increased sample or computational efficiency.

The aforementioned baselines are mixed and matched with the modeling choices (\cref{ssec:algo}) to form different baselines, e.g.\ MTL with TaskID (MTL-TaskID) or FineTuning with MH (FineTuning-MH). At the core of this work lies a particular combination, explained next.

\paragraph{Replay-based Recurrent RL (3RL)} 

A general approach to TACRL is to combine ER---one of CL's most versatile baselines---with an RNN, one of RL's most straightforward approaches to handling partial observability. 
We refer to this baseline as \emph{replay-based recurrent RL} (\algname{}). As an episode unfolds, \algname{}'s RNN representations $z_t = \text{RNN}(\{(s_i,a_i,r_i)\}_{i=1}^{t-1})$ should capture the specificities of the task ask, thus helping the actor $\pi_\theta(a|s,z)$ and critic $Q_\phi(s,a,z)$ in their respective approximations.
We will see in \Cref{sec:empirical}, that the RNN delivers more than expected: it enables forward transfer by decomposing new tasks and placing them in the context of previous ones.  
We provide pseudocode for \algname{} in \Algref{alg:3rl}. 
as in \cite{fakoor2019meta,Tianwei2021recurrent}, the actor and critics enjoy their own RNNs. They are thus parameterized by $\theta$ and $\phi$, respectively.
Our RNN implementation employs gated recurrent units (GRUs) \citep{Chung2014EmpiricalEO}, as prescribed by \cite{fakoor2019meta}. Similarly, ER and TX can be combined (ER-TX) into another baseline.

Let us re-examine the scenario involving a general-purpose robot in a continuous learning environment, where the robot incrementally acquires knowledge about manipulating new objects. Over time, it may become more resource-intensive to memorize the distinct manipulation strategies for each new object, as opposed to learning how to rapidly adapt to manipulate them. Consequently, a task-aware approach may lead to greater memory loss compared to a task-agnostic strategy. This line of thought gives rise to our second hypothesis.

\begin{minipage}{\linewidth}
\begin{algorithm}[H]
  \DontPrintSemicolon
  \footnotesize
  \KwEnv{a set of $K$ MPDs, allowed timesteps $T$}
  \KwInput{ initial parameters $\theta$, empty replay buffers $\mathcal{D}$ and $\mathcal{D}^{\text{old}}$ ,  replay cap $\beta$, batch size $b$, history length $h$}  
 
  \For{task $\tau$ in $K$}{
        set environment to $\tau^{th}$ MDP \\   
        \For{times-steps $t$ in $T$}{
            \tcc{Sampling stage}
            compute dynamic task representation $z_t = \text{RNN}_\theta(\{(s^o_i,a_i,r_i)\}_{i=t-h-1}^{t-1})$ \\
            observe state $s^o_t$ and execute action $a_t \sim \pi_\theta(\cdot|s^o_t, z_t)$ \\
            observe reward $r_{t}$ and next state $s^o_{t+1}$ \\
            store $(s^o_t, a_t, r_t, s^o_{t+1})$ in buffer $\mathcal{D}$ \\
            \tcc{Updating stage}
            sample a batch $B$ of $b \times min(\dfrac{1}{n}, 1-\beta)$ trajectories from the current replay buffer $\mathcal{D}$ \\ 
            append to $B$ a batch of $b \times min(\dfrac{n-1}{n}, \beta)$ trajectories  from the old buffer $\mathcal{D}^{\text{old}}$ \\
            Compute loss on $B$ and accordingly update parameters $\theta$ with one step of gradient descent \\
        }
        empty replay buffer $\mathcal{D}$ into $\mathcal{D}^{\text{old}}$
  }
  \caption{
  3RL in TACRL}
%   Pseudocode of 3RL in TACRL, \add{which we kept agnostic to the base algorithm and not tied to episodic RL.}}
  \label{alg:3rl}
\end{algorithm}
\end{minipage}

\begin{tcolorbox}[enhanced,attach boxed title to top center={yshift=-3mm,yshifttext=-1mm},
  colback=blue!5!white,colframe=blue!75!black,colbacktitle=red!80!black,
  title=Hypothesis 2,fonttitle=\bfseries,
  boxed title style={size=small,colframe=red!50!black} ]
    In the context of continual learning, the effect of Hypothesis 1 may be amplified due to the reduced impact of catastrophic forgetting in algorithms that continually learn to adapt, in contrast to algorithms that attempt to memorize each task.
\end{tcolorbox}

Next, we test both hypotheses, they could have significant implications for the practical application of continual reinforcement learning. 

%%%%%%%%%%%%%%%%%%%%%%%%%%%%
\section{Empirical Findings}
%%%%%%%%%%%%%%%%%%%%%%%%%%%%
\label{sec:empirical}

We now detail the empirical methodology to evaluate Hypotheses 1 (\Secref{sec:emp_hyp1}) and 2 (\Secref{sec:emp_hyp2}) using both synthetic data and challenging robotic manipulation tasks.

%---------------------
\paragraph{Benchmarks}

To evaluate our hypotheses, it is useful to have a mechanism for controlling the complexity of the available tasks. 
We propose a synthetic data benchmark that allows for the manipulation of the dimensions of observations and actions. 
The benchmark, which we refer to as Quadratic Optimization, involves maximizing a multidimensional quadratic function, with each task having its own set of parameters for $A_\tau$, $b_\tau$, and $c_\tau$. 
The reward function for each task is defined as the value of the resulting quadratic function  $r(s^o, \tau) = {s^o}^\top A_\tau s^o + b_\tau s^o + c_\tau$, where $s^o$ is the observation vector and $\tau$ is the task. 
We sample $A_\tau$ to be negative definite to ensure a global maximum for the quadratic function, and we set $c_\tau$ such that all tasks have the same maximum reward. 
The transition function is fixed and defined as $s_{t+1}^o = s^0_t + a_t$ where $a_t \in [-1,1]^d$ and $d$ is the dimentionality of the problem. 
Our benchmark provides a standardized platform for evaluating the performance of reinforcement learning algorithms in various task settings.

The second benchmark we study is Meta-World \citep{yu2019meta}, the canonical evaluation protocol for multi-task reinforcement learning (MTRL) \citep{yu2020gradient,yang2020multi,kumar2020discor,sodhani2021multi}. 
Meta-World offers a suite of 50 distinct robotic manipulation environments.
What differentiates Meta-World from previous MTRL and meta-reinforcement learning benchmarks \citep{rakelly2019efficient} is the broadness of its task distribution. 
Specifically, the different manipulation tasks are instantiated in the same state and action space\footnote{the fixed action space is an important distinction with traditional incremental supervised learning} and share a reward structure, i.e., the reward functions are combinations of reaching, grasping, and pushing different objects with varying shapes, joints, and connectivity. 
Meta-World is thus fertile ground for algorithms to transfer skills across tasks while representing the types of tasks likely relevant for real-world RL applications (see \Cref{fig:fig1} for a rendering of some of the environments) and as a result, its adoption in CRL is rapidly increasing \citep{ContinualWorld,mendez2020lifelong,Berseth2021CoMPSCM,wolczyk2022disentangling}.
Benchmarking in this space is also an active research area~\citep{mendez2022reuse}.

We utilize a subset of Meta-World called \texttt{CW10}, a benchmark introduced in \cite{ContinualWorld} with a particular focus on \emph{forward transfer}, which measures a method's ability to outperform one trained from scratch on new tasks. \texttt{CW10} comprises a specific subset of Meta-World tasks that are conducive to forward transfer and prescribes 1M steps per task, where a step corresponds to a sample collection and an update.
% We selected \codeword{CW10} over \codeword{CW20}, which repeats the \codeword{CW10} twice, to save compute and thus perform a more extensive analysis.

We introduce a new benchmark, referred to as \texttt{MW20}, which consists of the first 20 tasks of Meta-World, designed to explore a more challenging regime. The task sequence is twice as long, and data and computation are constrained to half, i.e., 500k steps per task. This benchmark allows us to investigate the performance and robustness of learning algorithms under more stringent conditions. We assembled all experiments using the Sequoia software for continual-learning research~\citep{normandin2022sequoia}.

For metrics, the global return/success and current return/success are the average success on all tasks and the average success on the task that the agent is currently learning, respectively.

%-------------------------------
\paragraph{Experimental Details}

In our synthetic data experiments, we manipulate several parameters to ensure controlled variation in scenario difficulty, including the number of tasks, observation and action dimensions, and the total number of timesteps. To account for variations in performance across different scenarios, we standardize performance for each combination of these variables. Consequently, the performance plots we present should be interpreted relative to one another, rather than as absolute measures of performance.

We conducted an exhaustive random search of over 80,000 runs, varying hyperparameters for all possible methods. 
To report robustness, we calculate the interquartile mean (IQM), meaning the top and bottom 25\% of runs are excluded, as prescribed in \cite{agarwal2021deep}. For maximal performance, we report the mean of the top 10\% of runs (note that the maximal performance results are reported only in the appendix).
The reported shaded areas are 2 standard errors. 
Full details of our experimental setup can be found in \appref{app:synthetic_data_details}.

As for the robotics tasks, we use the hyperparameters prescribed by Meta-World for their Multi-task SAC (MTL-SAC) method (see \appref{app:hyperparams_mw}).
We ensured the performance of our SAC implementation on the \codeword{MT10}, one of Meta-World's prescribed MTRL benchmarks, matches theirs (see \appref{app:MT10}).
For an explanation as to why 1) our reported  performances are  lower than those from the original Continual World, and 2) our baselines struggle with tasks learned easily in the single-task learning (STL) regime from the original Meta-World paper~\citep{yu2019meta}, we refer to \appref{app:mismatch}.
% For more details on our training procedures and hyperparameters, see \appref{app:exp_details}.
We test the methods using 8 seeds and report 90\% T-test confidence intervals as the shaded area of the figures.
As explained in \algref{alg:3rl}, we sometime oversample recently gathered data. 
When oversampling, we set the replay cap at 80\%, thus always spending at least 20\% of the compute budget on the current task. 
It is worth mentioning that for the robotic experiments, we opted not to include transformer-based baselines, as they were consistently outperformed by their RNN counterparts in the synthetic experiments.

%----------------------------------------------------
\subsection{Hypothesis 1: Can Fast Adaptation Overcome Tasks Memorization?}
\label{sec:emp_hyp1}

\begin{figure}
  \centering
    \includegraphics[width=0.33\linewidth]{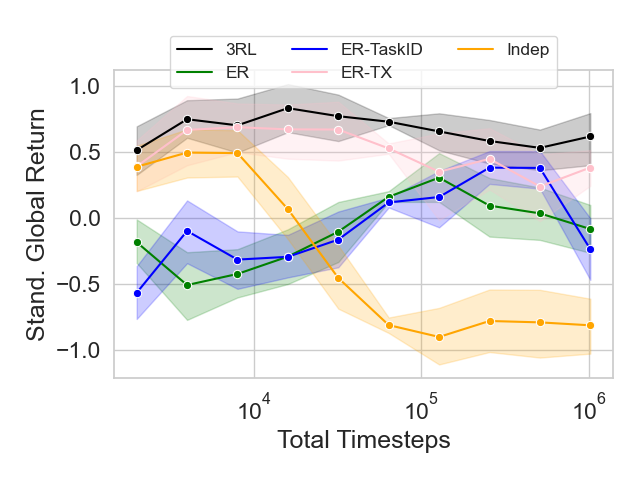} 
    \includegraphics[width=0.33\linewidth]{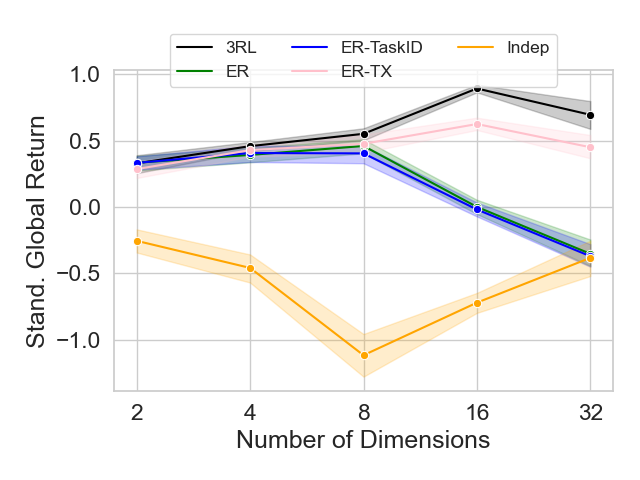}
    \includegraphics[width=0.33\linewidth]{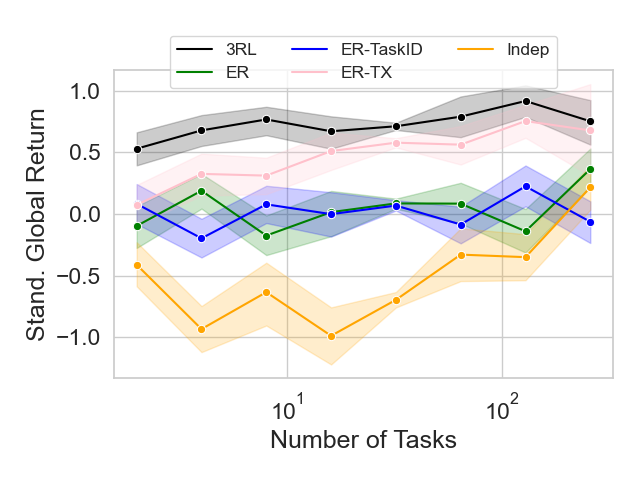}
    \caption{\small{
    \textbf{Standardized Results (IQM) for the CL Synthetic Tasks (40K Runs)}
        For each combination of studied dimensions (total timesteps, number of dimensions, number of tasks), we have standardized the performance of the runs, resulting in standardized runs. 
        Therefore, the plots reflect the ranking of the approach and not their absolute performance. We report the Interquartile mean of each method as a proxy for robustness. 
        As expected, \algname{} outperforms ER and TaskID in low timesteps regimes and ranks better as the number of observed dimensions increases. 
        However, the superior performance of the RNN is independent of the number of tasks considered. On the other hand, attention-based modeling is always outperformed or on par with recurrent modeling.
        The maximal performance plots and the plots with all methods are found in \appref{app:synthetic_top10_cl} and \appref{app:synthetic_cl_complete}.
    }}
    \label{fig:synthetic_plots_IQM}
\end{figure}

We present the initial results of our comprehensive synthetic empirical study in Figure \ref{fig:synthetic_plots_IQM}. We compare a subset of continual learning approaches across a range of scenario complexities. A clear pattern emerges, with \algname{} demonstrating superior robustness, particularly in regimes with higher dimensionality. This observation aligns with the intuition that as the dimensionality of a problem increases, it becomes more challenging to memorize all tasks (ER-TaskID) or find a single optimal solution for all tasks (ER), compared to learning a general solution that can adapt to specific tasks at test time (\algname{}). This finding is encouraging, as \algname{} has the potential to address the fragility of deep reinforcement learning in real-world settings \citep{dulac2019challenges}.

In terms of maximal performance (see \appref{app:synthetic_cl_complete}), we do not observe any significant patterns in the results. It appears that, for the Quadratic Optimization benchmark, identifying the appropriate hyperparameters is more crucial than the choice of method to achieve maximal performance. In addition, we provide a hyperparameter sensitivity analysis for the transformer-based approaches in \appref{app:tx_hparam_analysis}, where the type of positional encoding, learnable token embedding, the number of attention heads, and the transformer hidden state size are examined.

\begin{figure}[t]
    \begin{minipage}{0.56\linewidth}
    \centering
        \begin{subfigure}[b]{0.495\textwidth}
            \includegraphics[width=\linewidth]{figures/quad_opt/CorrMatrix_IQM-standard_AllRuns_All} 
            \caption{All runs}
            \label{fig:sub1}
        \end{subfigure}
        \hfill
        \begin{subfigure}[b]{0.495\textwidth}
            \includegraphics[width=\linewidth]{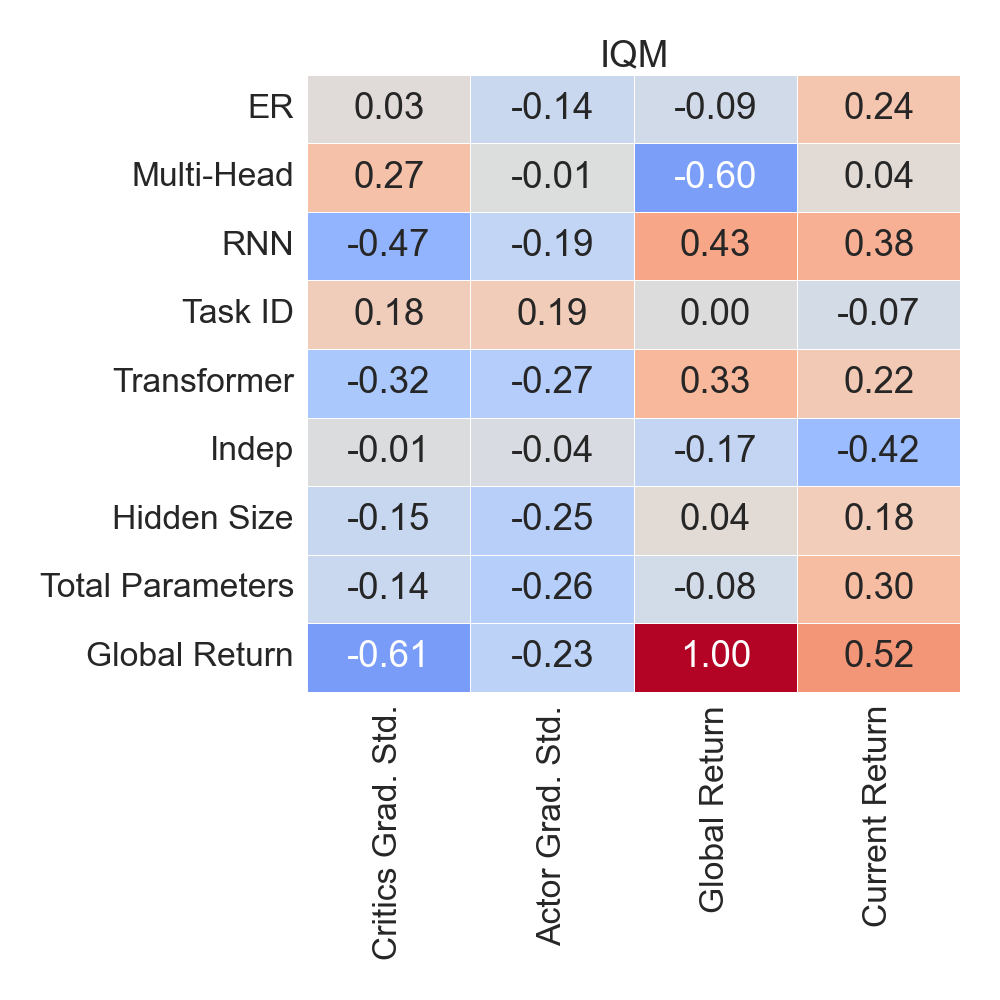}
            \caption{challenging CL scenario}
            \label{fig:sub2}
        \end{subfigure}
    \caption{\small{
        \textbf{Spearman Correlation Matrix in the of Synthetic Experiments (80k Runs).}
        The challenging CL scenario consists of 32 tasks, 32 dimensions, and 1M timesteps.
        Our analysis reveals that the RNN model enhances both robustness and maximal performance across studied slices, as indicated by the IQM (here) and top 10\% (\appref{app:synthetic_corr_matrices}) plots, respectively. 
        The RNN model proves especially helpful in the challenging CL scenario, where it potentially achieves superior performance through a reduction in gradient conflict. 
        Additionally, our findings show that the Task ID also improves robustness and maximal performance, although to a lesser extent and not through gradient conflict reduction. 
    }}
    \label{fig:corr_matrix}
    \end{minipage}
    \hspace{7pt}  
    \begin{minipage}{.42\linewidth}
        \includegraphics[width=\linewidth]{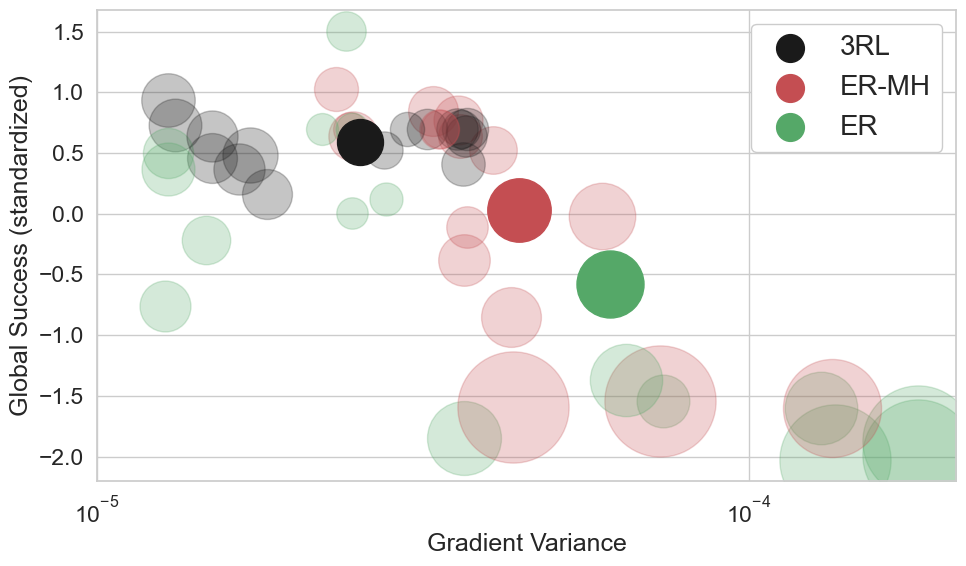} 
        \caption{\small{\textbf{In Meta-World, \algname{} decreases gradient conflict leading to an increase in training stability and performance.} The global success and gradient as measured by the variance of the gradients shown are plotted against each other. Training instability as measured by the variance of the Q-values throughout learning is represented by the markers' size, in a log scale. Transparent markers depict seeds, whereas opaque ones depict the means. We observe a negative correlation between performance and gradient conflict (-0.75) as well as performance and training stability (-0.81), both significant under a 5\% significance threshold. The hypothesis is that \algname{} improves performance by reducing gradient conflict via dynamic task representations. 
        }}
        \label{fig:stability}
    \end{minipage} 
\end{figure}

Motivated by the well-known fact that optimizing multiple tasks simultaneously can often result in the emergence of conflicting gradients or task interference \citep{yu2020gradient}, we  investigate whether such phenomena could  impede robustness and maximal performance in synthetic benchmarks. Furthermore, we conjecture that the RNN possesses an inherent capability to contextualize new tasks based on prior knowledge, thereby mitigating task interference and consequently exhibiting superior performance.

To this end, we focus on one of the most challenging scenarios of quadratic optimization which aligns with real-world use cases and our experiments in Meta-world. This scenario consists of 32 tasks, in which we test the ability to learn in the highest number of observation and action dimensions studied (32) and to sustain a long training period (1 million timesteps). Such a setting represents an important test for the different approaches, enabling us to gain further insights into the results and understand how and why \algname{} outperforms the other methods.

To test for this effect, we use the standard deviation of the gradients \emph{averaged over the dimension of the parameters} on the mini-batch throughout the training as a proxy of task interference or gradient conflict.
We explain in \appref{app:grad_variance} why we use the gradients' standard deviation to measure  gradient conflict instead of using the angle between the gradients as in \cite{yu2020gradient}. We compute correlation matrices between different method attributes and metrics as shown in \Cref{fig:corr_matrix}.
In the challenging continual learning scenario, we once again observe that the RNN approach consistently outperforms other methods in terms of both robustness and maximal performance, as evidenced by it achieving the highest correlation with global return.  

Consisten with our hypothesis, we observe that the RNN approach improves performance in the synthetic data benchmark by addressing gradient conflict, which negatively correlates with performance. This finding uncovers a mechanism through which the RNN approach enhances performance, especially in challenging scenarios. The RNN is followed by TX in both performance ranking and capability to mitigate gradient conflict. Additionally, our results indicate that this conflict might be a contributing factor to the poor performance of the multi-head approach in the synthetic data benchmark. More correlation matrices can be found in \appref{app:synthetic_corr_matrices}.

Our next experiments are performed in the domain of continual robotic manipulation tasks.
We postulate that the well-known deadly triad issue \citep{Sutton1998,Hasselt2018DeepRL}, which emerges from the combination of function approximation, bootstrapping, and off-policy learning, and potentially leads to divergence in value estimate, is a crucial problem in this context. The non-stationary task distribution, coupled with the intricate Meta-World environment, further exacerbates the phenomenon. 
If true, we anticipate that \algname{}, with its capacity to reduce gradient conflict, especially in lengthy and high-dimensional training regimes, may help improve stability and prove to be a superior CRL approach in more realistic tasks.

\begin{figure}
  \centering
    \includegraphics[width=0.45\linewidth]{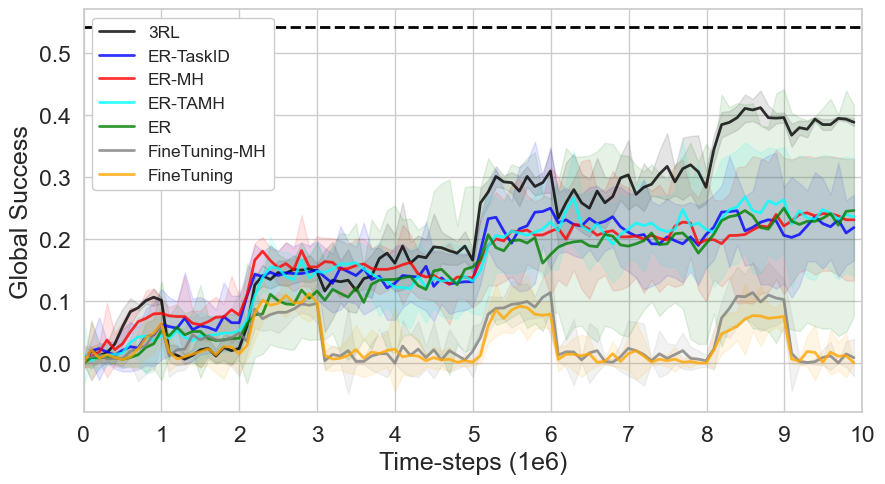} 
    \includegraphics[width=0.45\linewidth]{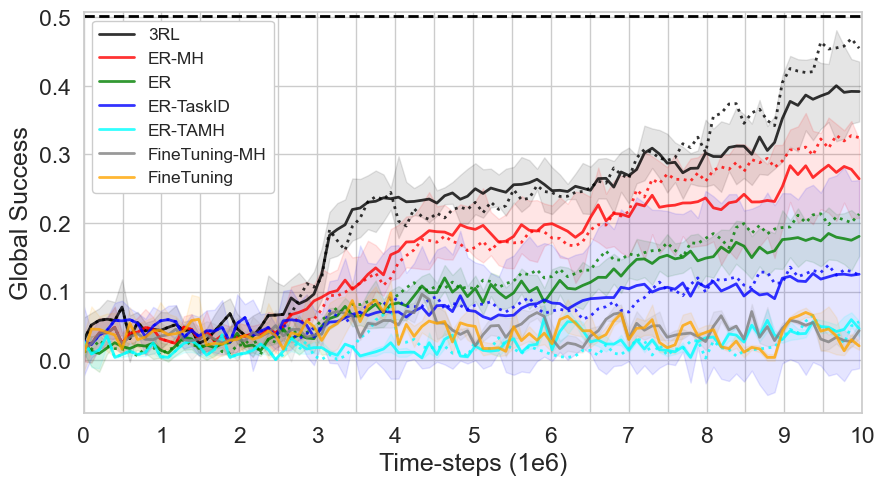}
    \caption{\small{\textbf{3RL outperforms all baselines in both \texttt{\small\textcolor{blue}{CW10}} (left) and \texttt{\textcolor{blue}{MW20}} (right)}. 
        The dashed horizontal black line at the top is the reference performance of training independent models on all tasks.  The dotted lines (right plot) represent the methods' performance when they oversample recently collected data. 
        \algname{} outperforms all other task-agnostic and more interestingly  task-aware baselines. 
        As a side note, ER is high variance as it attempts to solve the POMDP directly without having any explicit or implicit mechanism to do so.
    }}
    \label{fig:crl_cw10_mw20}
\end{figure}

Our empirical evaluation compares \algname{} to several baselines on the \codeword{CW10} and \codeword{MW20} benchmarks, as presented in \Cref{fig:crl_cw10_mw20}. In these results, \algname{} outperforms all other competing methods on both robotic benchmarks. This further supports our hypothesis that fast adaptation can outperform task memorization in challenging regimes.

To study the deadly triad hypothesis, we conduct an analysis of global performance, gradient conflict, and training stability, which is presented in \Cref{fig:stability}. 
Our findings provide empirical support for the hypothesis that one mechanism through which the RNN improves performance is indeed reduced gradient conflict and increased stability in the Q values.

To strengthen the conclusion reached in the study, several additional observations specific to the robotic benchmarks are provided in the Appendix. 
One of the experiments demonstrates that the difference in the number of parameters  of the methods is not responsible for the results obtained (\appref{app:larger_nets}). 
Another observation shows that combining task awareness with a recurrent neural network does not lead to improved performance (\appref{app:er-rnn-mh}).
Furthermore, we demonstrate that the RNN does not individually improve single-task performance (\appref{app:stl}).
The mechanism responsible for the superior performance achieved by the proposed algorithm is not parameter stability (\appref{app:param_stability}), i.e.\ the tendency of a parameter to stay within its initial value while new knowledge is incorporated.
Finally, some support is provided for the hypothesis that the RNN is able to place new tasks within the context of previous ones, thus facilitating forward transfer and improving optimization (\appref{app:recontext}). This is backed up by qualitative evidence presented in Figure 1.

%-----------------------------------------------------------
\subsection{Contrasting Continual Learning and Multi-task Learning}
\label{sec:emp_hyp2}

\begin{figure}[t]
\centering
    \begin{minipage}{.46\textwidth}
        \includegraphics[width=\linewidth]{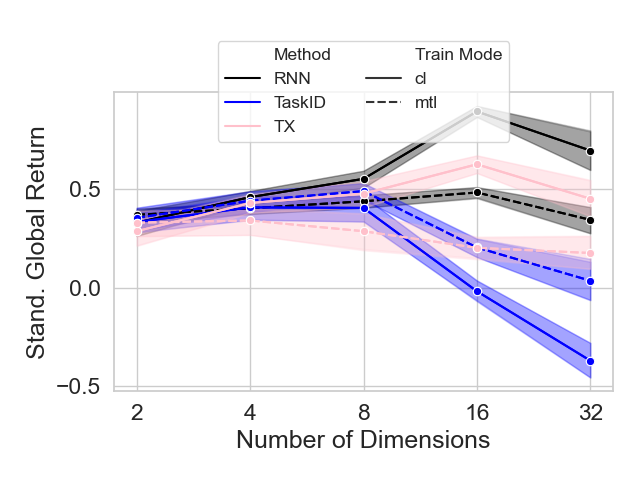}
        \caption{\small{
            \textbf{Standardized Results (IQM) for the CRL and MTRL Synthetic Tasks (80k Runs)}.
            Aligned with hypothesis \#2, \algname{}'s as well as ER-TX outperformances over tasks memorization is amplified in in continual learning as the complexity of the tasks increases and might be caused by fast adaptation.
        }}
    \label{fig:cl_mtl_obs_dim}
    \end{minipage}
    \hspace{7pt}  
    \begin{minipage}{.48\textwidth}
        \includegraphics[width=\linewidth]{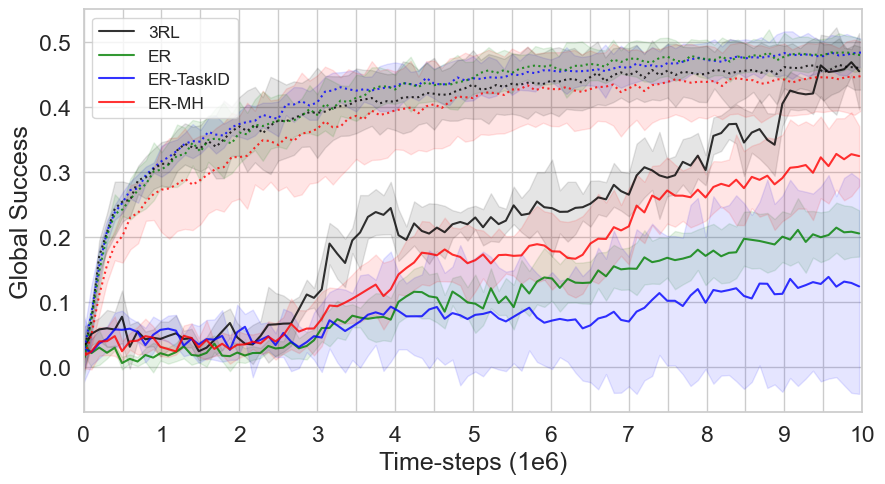} 
        \captionof{figure}{\small{\textbf{3RL reaches its MTRL soft-upper bound}.
          Continual vs multi-task learning methods in solid lines and dotted lines, respectively.
          \algname{} methods match its soft-upper bound MTL analog as well as the other MTRL baselines. 
          In contrast, other the performance of other baselines is drastically hindered by the non-stationary task distribution.
        }}
    \label{fig:mw20-500k}
    \end{minipage}  
\end{figure}

We proceed to investigate our Hypothesis \#2, which postulates that the advantages of fast adaptation over task memorization are magnified in a continual learning setting, due to the phenomenon of catastrophic forgetting. 
We conduct a comparative study between continual learning and multi-task learning methods and report our findings in \Cref{fig:cl_mtl_obs_dim}. 
As hypothesized, TaskID exhibits a significant decrease in relative performance with increasing dimensionality in CL. 
Additionally, we observe that \algname{} outperforms its multi-task learning counterpart in high-dimensional scenarios, which is a surprising finding since multi-task learning is often used as a soft upper bound in evaluating CL methods in both supervised~\citep{Aljundi2019Online, lopez2017gradient, delange2021continual} and reinforcement learning~\citep{rolnick2019experience,Traore19DisCoRL,ContinualWorld}.

We end by comparing CRL and MTRL approaches in the robotic environment.
In~\Cref{fig:mw20-500k} we report, for the second time, the results of the \codeword{MW20} experiments.
This time, we focus on methods that oversample the current task and more importantly, we report the performance of each method's multi-task analog, i.e.\ their soft-upper bound (dotted line of the same color).
Note that FineTuning methods do not have an MTRL counterpart and are thus not included in the current analysis.
\algname{} is the only approach that matches the performance of its MTRL equivalent.
We believe it is the first time that a specific method achieves the same performance in a non-stationary task regime compared to the stationary one, amidst the introduced challenges like forgetting.

%%%%%%%%%%%%%%%%%%%%%%
  \section{Related Work}
  %%%%%%%%%%%%%%%%%%%%%%
  \label{sec:related}

CRL is an area of research that has garnered significant interest and attention in recent years \citep{Kaplanis2018ContinualRL,schwarz2018progress,isele2018selective,kaplanis2020continual}. 
To study CRL in realistic settings, \cite{wolczyk2021continual} introduce the Continual World benchmark and discover that many CRL methods that reduce forgetting lose their transfer capabilities in the process, i.e.\ that policies learned from scratch generally learn new tasks faster than continual learners. 
Previous works study CL and compare task-agnostic methods to their upper bounds \citep{zeno2019task,van2019three} as well as CL methods compare to their multi-task upper bound \citep{ribeiro2019multi, rolnick2019experience, ammar2014online}. 
Refer to \cite{khetarpal2020towards} for an in-depth review of continual RL as well as \cite{lesort2021understanding,hadsell2020embracing} for a CL in general.

Closer to our training regimes, TACRL is an actively studied field. 
\cite{Xu2020TaskAgnosticOR} uses an infinite mixture of Gaussian Processes to learn a task-agnostic policy.
\cite{Kessler2021SameSD} learns multiple policies and casts policy retrieval
as a multi-arm bandit problem.
As for \cite{Berseth2021CoMPSCM,nagabandi2018deep}, they use meta-learning to tackle the task-agnosticism part of the problem.  

RNNs were used in the context of continual supervised learning in the context of language modeling \citep{Wolf2018ContinuousLI, wu2021pretrained} as well as in audio \citep{ehret2020continual, wang2019continual}. 
We refer to \cite{cossu2021continual} for an in-depth review of RNN in continual supervised learning. 

As in our work, RNN models have been effectively used as policy networks for reinforcement learning, especially in POMDPs where they can effectively aggregate information about states encountered over time \citep{wierstra2007solving, fakoor2019meta,Tianwei2021recurrent,heess2015memory}. Recently, transformers have also gained popularity in RL for modeling sequences \citep{chen2021decision,janner2021offline}.
RNNs were used in the context of MTRL \cite{8851824}.
Closer to our work, \cite{sorokin2019continual} leverages RNNs in a task-aware way to tackle a continual RL problem. 
To the best of our knowledge, RNNs have not been employed within TACRL nor combined with Experience Replay in the context of CRL. 

%%%%%%%%%%%%%%%%%%%%
\section{Conclusion}
%%%%%%%%%%%%%%%%%%%%
\label{sec:discuss}

In this work, we seek to understand the performance differences between task-agnostic continual learning (CL) and multi-task (MTL) agents, focusing on the potential benefits of task-agnosticity. We propose two hypotheses and introduce a replay-based recurrent reinforcement learning (\algname{}) method to evaluate them.

We conduct experiments using synthetic tasks and the Meta-World benchmark, examining settings with limited resources, high dimensionality, and numerous tasks. Our results demonstrate that \algname{} consistently outperforms baseline methods and, in some cases, even surpasses its multi-task equivalent. This performance improvement is partly attributable to a decrease in gradient conflict.

The findings challenge conventional assumptions about the inherent challenges of task-agnostic CL. They suggest that methods like \algname{} may offer significant advantages in resource-constrained, high-dimensional, and multi-task environments, highlighting the potential of task-agnostic CL over task-aware MTL approaches.

\subsubsection*{Acknowledgments}

We would like to thank Timothée Lesort and Lucas Caccia for providing valuable feedback.
We also thank Samsung Electronics Co., Ldt. for their support.

\bibliography{references}
\bibliographystyle{collas2023_conference}

\clearpage
\appendix
\begin{center}
  {\LARGE  \bf Appendix: \ 
  Task-Agnostic Continual Reinforcement Learning: Gaining Insights and Overcoming Challenges
  }
  \end{center}
  
  %%%%%%%%%%%%%%%%%%%%%%%%%%%%%%
  \section{Extending TACRL's related settings}
  %%%%%%%%%%%%%%%%%%%%%%%%%%%%%%
  \label{app:pomdp_further}
  
  We continue the discussion on TACRL's related settings. 
  In Meta-RL \citep{finn2017model}, the training task, or analogously the training hidden-states $s^h$, are different from the testing ones. 
  We thus separate them into disjoint variable $s^h_{\text{train}}$ and $s^h_{\text{test}}$.
  In this setting, some fast adaptation to $s^h$ is always required.
  Meta-RL does not deal with a non-stationary training task distribution. 
  Its continual counterpart however, i.e. Continual Meta-RL~\citep{Berseth2021CoMPSCM}, does.
  \tblref{tab:ext_tacrl} summarizes the settings. 
  
  Noteworthy, the Hidden Parameter MDP (HiP-MDP)~\citep{DoshiVelez2016HiddenPM} is similar to the HM-MDP but assumes a hidden states, i.e., the hidden states are samples i.i.d. 
  Another setting similar to the HM-MD is the Dynamic Parameter MDP \citep{xie2020deep} in which the hidden-state are non-stationary but change at every episode.  
  
  \begin{center}
  \tabcolsep=2pt\relax
   \scriptsize%
   \makebox[\textwidth]{\begin{tabular}{cccccc}
          & T  & $\pi$ & Objective & Evaluation \\
          \hline \hline \\
          MDP \citep{Sutton1998} & $p(s_{t+1} | s_t, a_t)$ & $\pi(a_t | s_t)$ & $\underset{\pi}{\EE} \big[\sum_{t=0}^\infty \gamma^{t} r_t \big]$ & - \\
          \rowcolor{LightCyan}
          POMDP \citep{kaelbling1998planning} & $p(\blue{s_{t+1}^h, s_{t+1}^o} | \blue{s_t^h, s_t^o}, a_t)$ & $\pi(a_t | \blue{s^o_{1:t}, a_{1:t-1}, r_{1:t-1}})$ & $\underset{\blue{s^h}}{\EE} \Big[ \underset{\pi}{\EE} \big[\sum_{t=0}^\infty \gamma^{t} r_t \big] | \blue{s^h} \Big]$ & - \\
          HM-MDP \citep{choi2000hidden} & $ p(s_{t+1}^o | \blue{s_{t+1}^h}, s_t^o, a_t) p(\blue{s_{t+1}^h}
          | s_t^h)$ & $\pi(a_t | s^o_{1:t}, a_{1:t-1}, r_{1:t-1})$ & $\underset{s^h}{\EE} \Big[ \underset{\pi}{\EE} \big[\sum_{t=0}^\infty \gamma^{t} r_t \big] | s^h \Big]$ & - \\
          \rowcolor{LightCyan}
          \textbf{Task-agnostic CRL} & $ p(s_{t+1}^o | s_{t+1}^h, s_t^o, a_t) p(s_{t+1}^h | s_t^h)$ & $\pi(a_t | s^o_{1:t}, a_{1:t-1}, r_{1:t-1})$ & $\underset{s^h}{\EE} \Big[ \underset{\pi}{\EE} \big[\sum_{t=0}^\infty \gamma^{t} r_t \big] | s^h \Big]$ & $\underset{\blue{\tilde{s}^h}}{\EE} \Big[ \underset{\pi}{\EE} \big[\sum_{t=0}^\infty \gamma^{t} r_t \big] | s^h \Big]$ \\
          Task-Aware CRL & $ p(s_{t+1}^o | s_{t+1}^h, s_t^o, a_t) p(s_{t+1}^h | s_t^h)$ & $\pi(a_t | \blue{s^h_t, s^o_t})$ & $\underset{s^h}{\EE} \Big[ \underset{\pi}{\EE} \big[\sum_{t=0}^\infty \gamma^{t} r_t \big] | s^h \Big]$ & $\underset{\tilde{s}^h}{\EE} \Big[ \underset{\pi}{\EE} \big[\sum_{t=0}^\infty \gamma^{t} r_t \big] | s^h \Big]$ \\
          \rowcolor{LightCyan}
          Multi-task RL & $ p(s_{t+1}^o | s_{t+1}^h, s_t^o, a_t) p(\blue{s_{t+1}^h})$ & $\pi(a_t | s^h_t, s^o_t)$ & $\underset{\blue{\tilde{s}^h}}{\EE} \Big[ \underset{\pi}{\EE} \big[\sum_{t=0}^\infty \gamma^{t} r_t \big] | s^h \Big]$ & - \\
          Meta RL \citep{finn2017model} & $ p(s_{t+1}^o | s_{t+1}^h, s_t^o, a_t) p(s_{t+1}^h)$ & $\pi(a_t | s^o_{1:t}, a_{1:t-1}, r_{1:t-1})$ & $\underset{\blue{\tilde{s}^h_{\text{train}}}}{\EE} \Big[ \underset{\pi}{\EE} \big[\sum_{t=0}^\infty \gamma^{t} r_t \big] | s^h_{\text{train}} \Big]$ & $\underset{\blue{\tilde{s}^h_{\text{test}}}}{\EE} \Big[ \underset{\pi}{\EE} \big[\sum_{t=0}^\infty \gamma^{t} r_t \big] | s^h_{\text{test}} \Big]$ \\
          \rowcolor{LightCyan}
          Continual Meta-RL & $ p(s_{t+1}^o | \blue{s_{t+1}^h}, s_t^o, a_t) p(\blue{s_{t+1}^h}
          | s_t^h)$ & $\pi(a_t | s^o_{1:t}, a_{1:t-1}, r_{1:t-1})$ & $\underset{s^h_{\text{train}}}{\EE} \Big[ \underset{\pi}{\EE} \big[\sum_{t=0}^\infty \gamma^{t} r_t \big] | s^h_\text{train} \Big]$ & $\underset{\blue{\tilde{s}^h_{\text{test}}}}{\EE} \Big[ \underset{\pi}{\EE} \big[\sum_{t=0}^\infty \gamma^{t} r_t \big] | s^h_{\text{test}} \Big]$ \\
              
      \end{tabular}}
      \captionof{table}{Summarizing table of the settings relevant to TACRL. For readability purposes, $\tilde{s}^h$ denotes the stationary distribution of $s^h$. The Evaluation column if left blank when it is equivalent to the Objective one.}
      \label{tab:ext_tacrl}
  \end{center}
  
  %%%%%%%%%%%%%%%%%%%%%%%%%%%
  \section{Soft-Actor Critic}
  %%%%%%%%%%%%%%%%%%%%%%%%%%%
  \label{app:sac}
  
  The Soft Actor-Critic (SAC)~\citep{sachaarnoja18b} is an off-policy actor-critic algorithm for continuous actions. SAC adopts a maximum entropy framework that learns a stochastic policy which not only maximizes the expected return but also encourages the policy to contain some randomness. 
  To accomplish this, SAC utilizes an actor/policy network (i.e. $\pi_\phi$) and critic/Q network (i.e. $Q_\theta)$, parameterized by $\phi$ and $\theta$ respectively. 
  Q-values are learnt by minimizing one-step temporal difference (TD) error by sampling previously collected data from the replay buffer~\citep{lin1992self} denoted by $\mathcal{D}$. 
  
  \begin{align}
  \label{eq:sac_q}
      \mathcal{J}_Q(\theta) = \underset{s, a}{\EE}  \Big[ \Big( Q_{\theta}(s, a) - y(s,a) \Big)^2 \Big], ~ a' \sim \pi_\phi(\cdot|s')
  \end{align}
  where $y(s,a)$ is defined as follows:  
  \begin{align*}
      &y(s,a) = r(s,a) + \gamma \underset{s', a'}{\EE} \Big [Q_{\hat{\theta}}(s', a') - \alpha \log(a'|s') \Big]
  \end{align*}
  
  And then, the policy is updated by maximizing the likelihood of actions with higher Q-values:
  \begin{align}
      \label{eq:sac_pi}
      \mathcal{J}_\pi(\phi) =  \underset{s, \hat{a}} \EE \Big[ Q_{\theta}(s, \hat{a}) - \alpha \log \pi_\phi(\hat{a}|s) \Big], ~ \hat{a} \sim \pi_{\phi}(\cdot|s)
  \end{align}
  where $(s,a, s') ~\sim \mathcal{D}$ (in both \eqref{eq:sac_q} and \eqref{eq:sac_pi}) and $\alpha$ is entropy coefficient. Note that although SAC is used in this paper, other off-policy methods for continuous control can be equally utilized for CRL. SAC is selected here as it has a straightforward implementation and few hyper-parameters. 
  
  %%%%%%%%%%%%%%%%%%%%%%%%%%%%%%%
  \section{Baselines Definitions}
  %%%%%%%%%%%%%%%%%%%%%%%%%%%%%%
  \label{app:ext_baselines}

  \textbf{FineTuning-MH} is FineTuning with task-specific heads. 
  For each new task, it spawns and attaches an additional output head to the actor and critics. 
  Since each head is trained on a single task, this baseline allows to decompose forgetting happening in the representation of the model (trunk) compared to forgetting in the last prediction layer (head). 
  It is a task-aware method.
  
  \textbf{ER-TaskID} is a variant of ER that is provided with task labels as inputs (i.e.\ each observation also contains a task label). 
  It is a task-aware method that has the ability to learn a task representation in the first layer(s) of the model. 
  
  \textbf{ER-MH} is ER strategy which spawns tasks-specific heads \citep{ContinualWorld}, similar to FineTuning-MH. ER-MH is often the hardest to beat task-aware baseline \citep{}. 
  ER-TaskID and ER-MH use two different strategies for modelling task labels. Whereas, MH uses $|h| \times |A|$ task-specific parameters (head) taskID only uses $|h|$, with $|h|$ the size of the network's hidden space (assuming the hidden spaces at each layer have the same size) and $|A|$ the number of actions available to the agent.  
  
  \textbf{ER-TAMH (task-agnostic multi-head)} is similar to ER-MH, but the task-specific prediction heads are chosen in a task-agnostic way. 
  Specifically, the number of heads is fixed a priori (in the experiments we fix it to the number of total tasks) and the most confident actor head, w.r.t. the entropy of the policy, and most optimistic critic head are chosen.
  ER-TAMH has the potential to outperform ER, another task-agnostic baseline, if it can correctly infer the tasks from the observations.

  \textbf{MTL} is our backbone algorithm, namely SAC, trained via multi-task learning.
  It is the analog of ER.
  
  \textbf{MTL-TaskID} is MTL, but the task label is provided to the actor and critic.
  It is the analog of ER-TaskID and is a standard method, e.g.\ \citep{haarnoja2018soft}.
  
  \textbf{MTL-MH} is MTL with a task-specific prediction network.
  It is the analog of ER-MH and is also standard, e.g.\ \citep{yu2020gradient, haarnoja2018soft, yu2019meta}.
  
  \textbf{MTL-TAMH} is similar to MTL-MH, but the task-specific prediction heads are chosen in the same way as in ER-TAMH.
  
  \textbf{MTL-RNN} is similar to MTL, but the actor and critic are mounted with an RNN. 
  It is the analog of \algname.
  
  \textbf{MTL-TX} is similar to MTL, but the actor and critic are mounted with a Transformer.

  %%%%%%%%%%%%%%%%%%%%%%%%%%%%%%
  \section{Experimental Details}
  %%%%%%%%%%%%%%%%%%%%%%%%%%%%%%
  \label{app:exp_details}
  
   %-------------------------------
  \subsection{Synthetic Data Hyperparameters}
  \label{app:synthetic_data_details}
  
  In \tblref{tab:synthetic_hyperparams} we report the hyperparameters ranges used for the synthetic benchmark experiments.
  
  \begin{table}[]
      \centering
      \begin{tabular}{l || l}
           \textbf{Setting} & \\
           total timesteps & [2,000, 4,000, 8,000, 16,000, 32,000, 64,000, 128,000, 256,000, 512,000, 1,024,000] \\
           number of tasks & [2, 4, 8, 16, 32, 64, 128, 256] \\
           number of dimensions & [2, 4, 8, 16, 32] \\
           CL or MTL & [CL, MTL] \\
           episode length & 100 \\
            \\
           \textbf{All Methods} & \\ 
           Architecture  & 2-layer MLP  \\
           activation & ReLU \\
           soft-target interpolation & $5 \times 10^{-3}$ \\
           learning rate & log-uniform(min=0.00001, max=0.1) \\
           batch size & uniform(min=2, max=256) \\
           warm-up period & uniform(min=2, max=100) \\
           burn-in period & uniform(min=2, max=100) \\
           hidden size & [8, 16, 32, 64, 128] \\
           automatic entropy tuning & [on, off] \\
           \\
           \textbf{ER methods} & \\
           replay cap $\beta$ & 1.0, 0.8, 0.5 \\
           buffer size & [1,000, 10,000, 100,000, 1,000,000] \\
            \\
            \textbf{methods with context (taskID, RNN, TX)} \\        
            context size & 30 \\
           \\
            \textbf{methods with RNN} \\
            number of layers & 1 \\
            history length & [2, 4, 8, 16] \\
           \\
            \textbf{methods with Transformers} \\     
            hidden size & [8, 16] \\
            number of heads & [1, 2] \\
            positional embeddings & [learned, sinusoidal] \\
            token  embeddings & [learned, None] \\
            history length & [2, 4, 8, 16] \\
      \end{tabular}
      \caption{\textbf{Table of hyperparameters for the synthetic benchmark hyperparameter search.}}
      \label{tab:synthetic_hyperparams}
  \end{table}

  %-------------------------------
  \subsection{Meta-World Hyperparameters and their justification}
  \label{app:hyperparams_mw}

  The choice of hyperparameters required quite some work. 
  Initially, we used the SAC hyperparameters prescribed by Continual World \citep{ContinualWorld}, designed on Meta-World v1, without any success. 
  We then tried some of Meta-World v2's prescribed hyperparameter, which helped us match MetaWorld’s multi-task reported results. 
  
  However, the continual and multi-task learning baselines would still suffer from largely unstable training due to the deadly triad \citep{vanhasselt2018deep} problems in CRL and MTRL.
  After further experimentation, we observed that gradient clipping could stabilize training, and that clipping the gradients to a norm of 1 achieved the desired behavior across all methods, except for the multi-head baseline in which 10 was more appropriate. 
  
  Lastly, we use automatic entropy tuning except in the MTRL experiments, where we found its omission to be detrimental. 
  Because their MT-SAC implementation learns a task-specific entropy term, we think this is the reason why \cite{yu2019meta} do not observe the same behavior.
  All hyperparameters are summarized in \tblref{tab:hyperparams}.
  
  \begin{table}[]
      \centering
      \begin{tabular}{l || c c c}
           Architecture  & 2-layer MLP  \\
           hidden state & [400, 400]  \\
           activation & ReLU \\
           episode length & Continual World: 200, else: 500 \\
           minimum buffer size & 10 tasks: 1500, 20 tasks: 7500 \\
           batch size & \\
           learning rate &  $1 \times 10^{-3}$ \\
           soft-target interpolation & $5 \times 10^{-3}$ \\
           burn in period & 10,000 steps \\
           RNN's number of layers & 1 \\
           RNN's context size & 30 \\
           RNN's context length & 15 \\
          \\
           & \textbf{Independent} & \textbf{MTRL} & \textbf{CRL} \\
           \hline
           automatic-entropy tuning & on & off & on \\
           gradient clipping & None & MH: 10, else: 1 & MH: 10, else: 1 \\
           & 
      \end{tabular}
      \caption{\textbf{Table of hyperparameters.} 
      The top hyperparameters are global, whereas the bottom ones are setting specific.}
      \label{tab:hyperparams}
  \end{table}

  %-------------------------------
  \subsection{Computing Resources}
  \label{app:comp_ressources}

For the synthetic benchmark, we ran 82,160 different runs. 
We used 2 or 4 CPU per runs, resulting in 10,260 days worth of compute.

For the Meta-world benchmarks, all experiments were performed on Amazon EC2's P2 instances which incorporates up to 16 NVIDIA Tesla K80 Accelerators and is equipped with Intel Xeon 2.30GHz cpu family. All meta-world experiments included in the paper can be reproduced by running 43 method/setting configurations with 8 seeds, each running for 4.2 days on average.
 
  %\textcolor{red}{For the entirety of this work, we have 16,430 GPU days of compute.}
  
  % official results 13: 1759 days
  % official results 14: 260 days
  % official results 15: 1409 days
  
  % mtl0: 1386 days
  % mtl2: 812 days
  
  % official results 15-2: 715 days
  % official results 16: 426 days
  % official results 17: 540 days
  
  % official results 18: 4708 days
  % official results 19: 2326 days
  % official results 20: 1110 days
  % official results 21: 841 days
  % official results 22: 138 days
 
%-------------------------------
\subsection{Resolving the confusion about what could seem as a result mismatch with other Meta-World experiments in the literature}
\label{app:mismatch}

The astute reader will find that:  1) our reported  performances are  lower than those from the original Continual World, and 2) our baselines struggle with tasks learned easily in the single-task learning (STL) regime from the original Meta-World paper \cite{yu2019meta}.
These discrepancies are explained by the fact that \citet{ContinualWorld} carried out their original  Continual-World study using Meta-World v1, which is far from the updated v2 version we use here. For example, the state space is thrice larger in v2 and the reward functions have been completely rewritten. 
The original Meta-World study of \citet{yu2019meta} was performed under far more generous data and compute settings (see \appref{app:MT10}) and relied on MTL and STL specific architectures and hyperparameters.

  %-------------------------------
  \subsection{Software and Libraries}
  \label{app:soft_lib}
  
  In the codebase we've used to run the experiments, we have leverage some important libraries and software.
  We used Mujoco~\citep{mujoco} and Meta-World \citep{yu2019meta} to run the benchmarks.
  We used Sequoia~\citep{normandin2022sequoia} to assemble the particular CRL benchmarks, including \codeword{CW10}.
  We used Pytorch~\citep{paszke2019pytorch} to design the neural networks.

%%%%%%%%%%%%%%%%%
\section{Synthetic Data Experiments}
%%%%%%%%%%%%%%%%%

\subsection{Top 10\% of runs in the CL synthetic data benchmark for the methods presented in the main section}
\label{app:synthetic_top10_cl}

\begin{figure}
  \centering
    \includegraphics[width=0.32\linewidth]{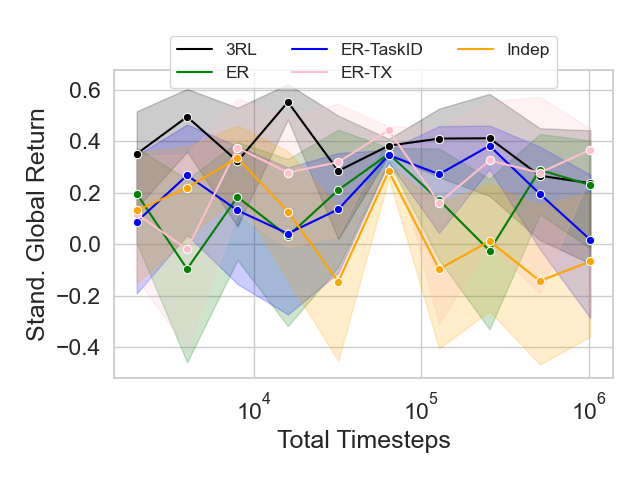}
    \includegraphics[width=0.32\linewidth]{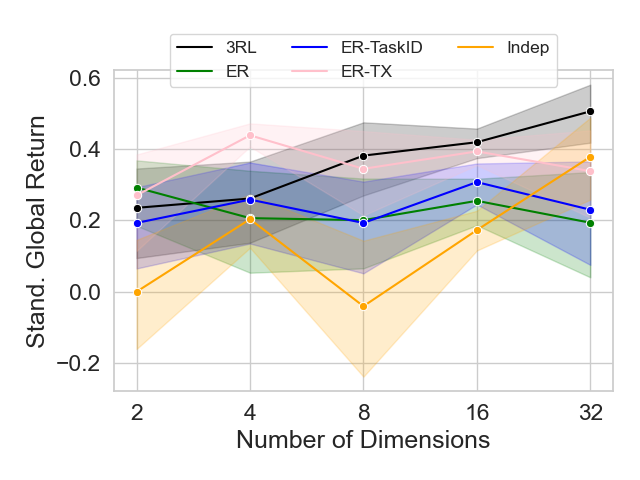}
    \includegraphics[width=0.32\linewidth]{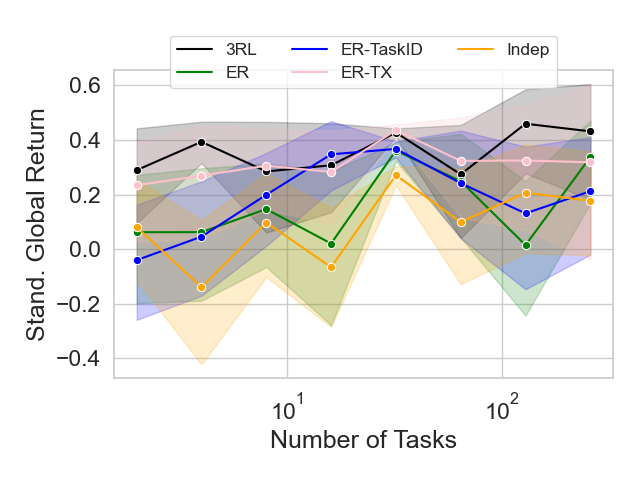}  
    \caption{\small{
        \textbf{standardized results (top 10\% of the runs) for the CL synthetic tasks (40k runs)} 
        Unlike for robustness, the impact of hyperpameter choice seems to overshadow the methods' impact on maximal performance.
    }}
    \label{fig:synthetic_plots_topk}
\end{figure}

See \figref{fig:synthetic_plots_topk} for our proxy of maximal performance in the CL synthetic benchmark, i.e. the average of the top 10\% runs.

\subsection{Complete results for the CL synthetic data experiments}
\label{app:synthetic_cl_complete}

\begin{figure}[h]
\centering
    \begin{subfigure}[b]{0.49\textwidth}
        \includegraphics[width=\linewidth]{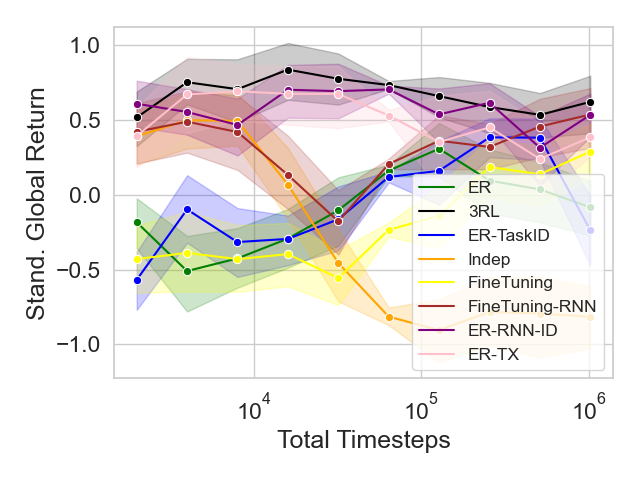} 
        \label{fig:sub1}
    \end{subfigure}
    \hfill
    \begin{subfigure}[b]{0.49\textwidth}
        \includegraphics[width=\linewidth]{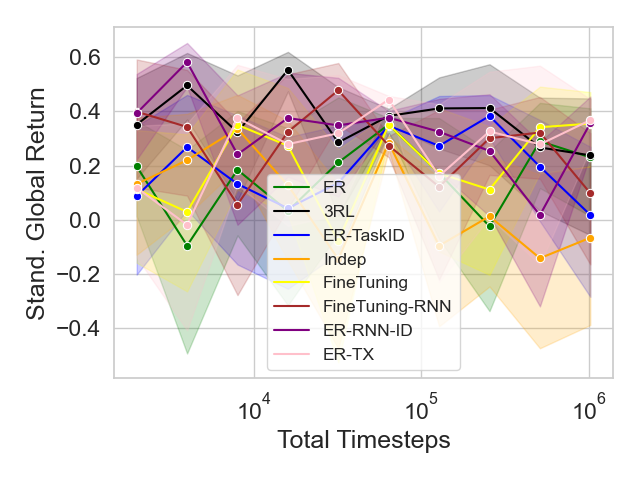} 
        \label{fig:sub2}
    \end{subfigure}
    
    \medskip
    
    \begin{subfigure}[b]{0.49\textwidth}
        \includegraphics[width=\linewidth]{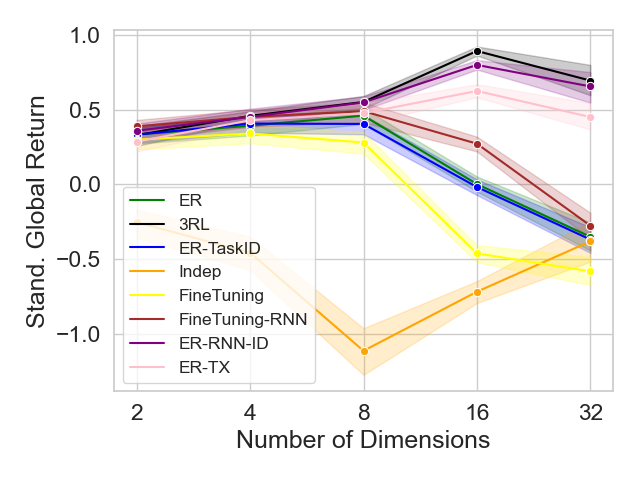} 
        \label{fig:sub1}
    \end{subfigure}
    \hfill
    \begin{subfigure}[b]{0.49\textwidth}
        \includegraphics[width=\linewidth]{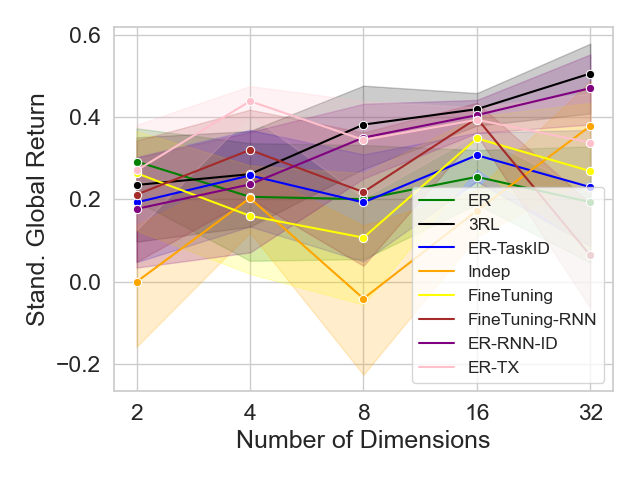} 
        \label{fig:sub2}
    \end{subfigure}
    
    \medskip
    
    \begin{subfigure}[b]{0.49\textwidth}
        \includegraphics[width=\linewidth]{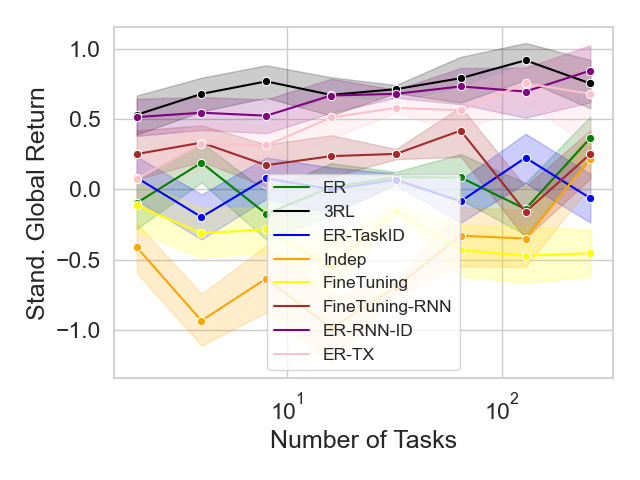} 
        \label{fig:sub1}
    \end{subfigure}
    \hfill
    \begin{subfigure}[b]{0.49\textwidth}
        \includegraphics[width=\linewidth]{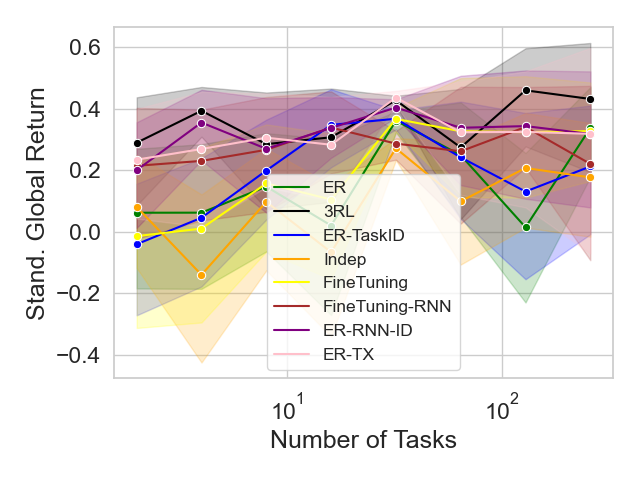} 
        \label{fig:sub2}
    \end{subfigure}
    
\caption{\small{
    \textbf{Complete results for the CL synthetic experiments (40k runs)}
    For readability, we removed the multi-head (MH) methods from the plots, as they are performing relatively too low to others. See \figref{fig:synthetic_cl_mtl_complete} for some MH results. On the left are IQM plots whereas on the right top 10\% ones.
}}
\label{fig:synthetic_cl_complete}
\end{figure}

See \figref{fig:synthetic_cl_complete} for the complete CL synthetic data experiments, i.e. when all methods are included.

\subsection{Hyperparameter analysis of the transformer-based approaches}
\label{app:tx_hparam_analysis}

\begin{figure}[h]
\centering
    
    \begin{subfigure}[b]{0.45\textwidth}
        \includegraphics[width=\linewidth]{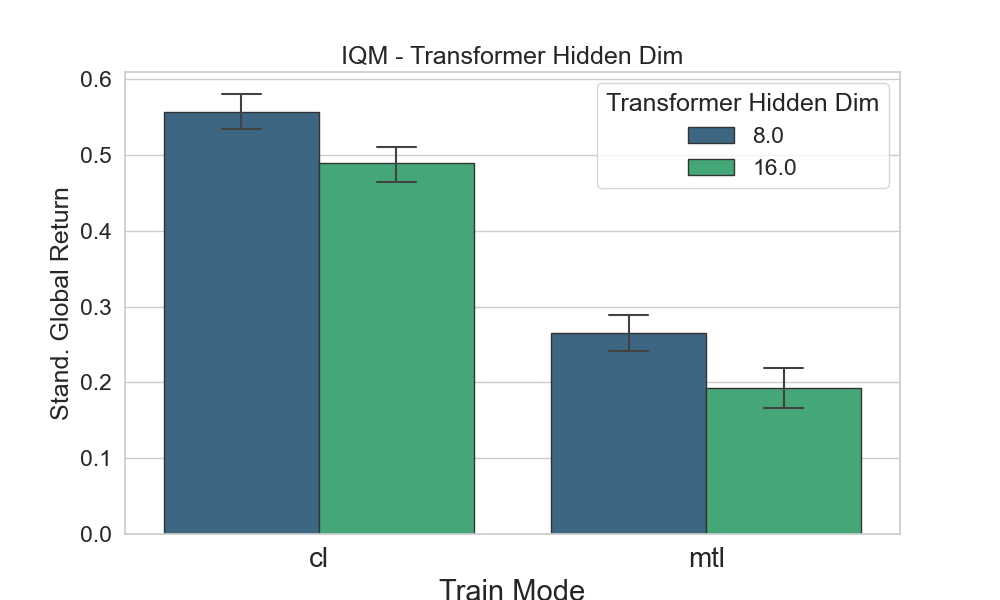} 
        \label{fig:sub1}
    \end{subfigure}
    \hfill
    \begin{subfigure}[b]{0.45\textwidth}
        \includegraphics[width=\linewidth]{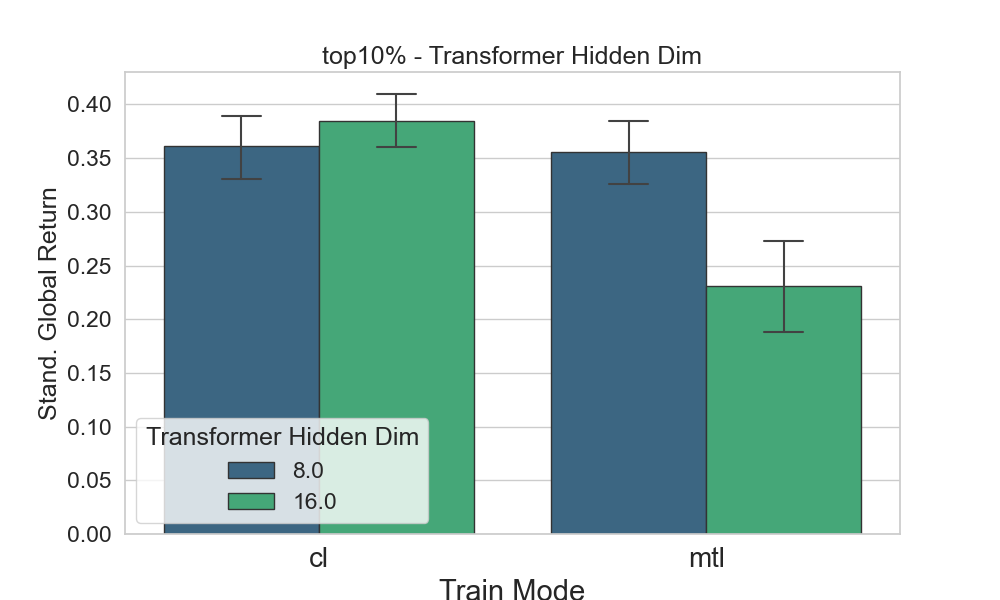} 
        \label{fig:sub2}
    \end{subfigure}
    
    \medskip

    \begin{subfigure}[b]{0.45\textwidth}
        \includegraphics[width=\linewidth]{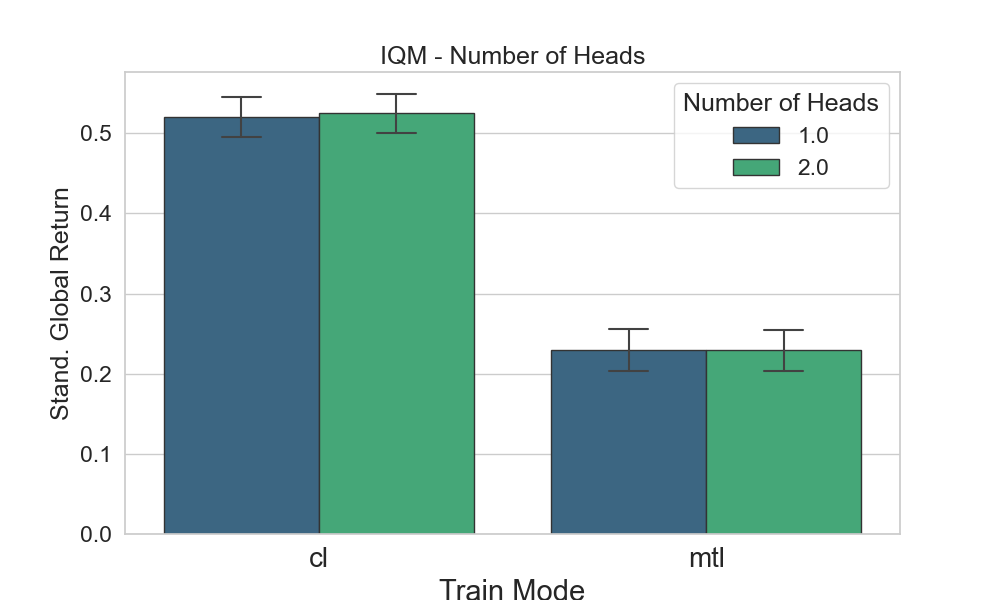} 
        \label{fig:sub1}
    \end{subfigure}
    \hfill
    \begin{subfigure}[b]{0.45\textwidth}
        \includegraphics[width=\linewidth]{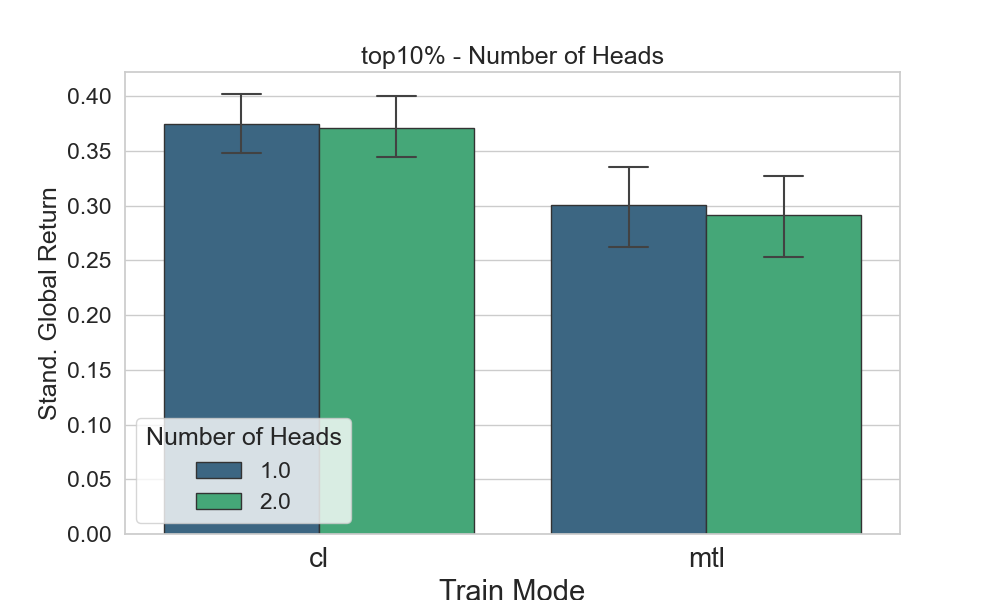} 
        \label{fig:sub2}
    \end{subfigure}
    
    \medskip
    
    \begin{subfigure}[b]{0.45\textwidth}
        \includegraphics[width=\linewidth]{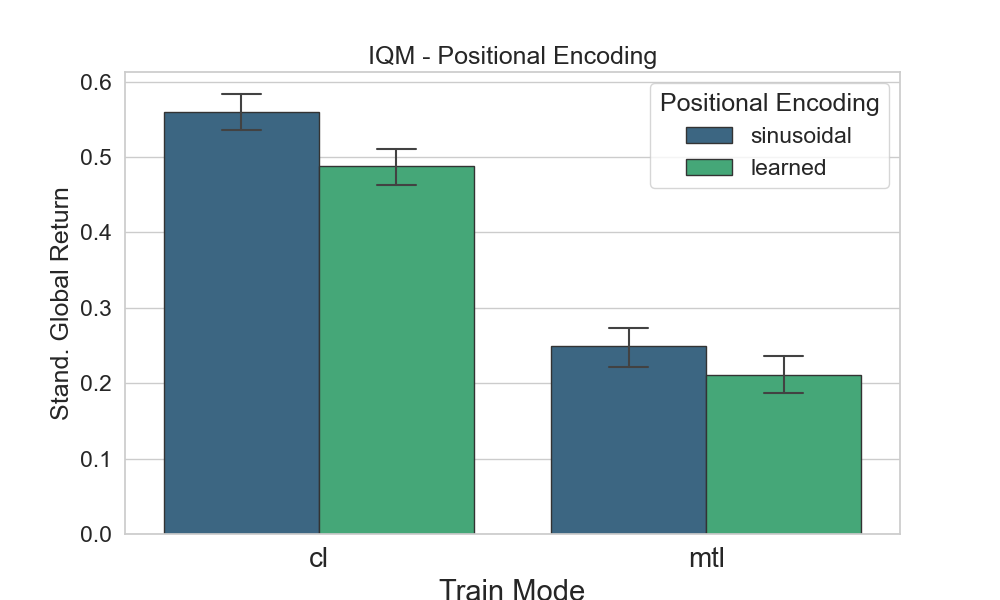} 
        \label{fig:sub1}
    \end{subfigure}
    \hfill
    \begin{subfigure}[b]{0.45\textwidth}
        \includegraphics[width=\linewidth]{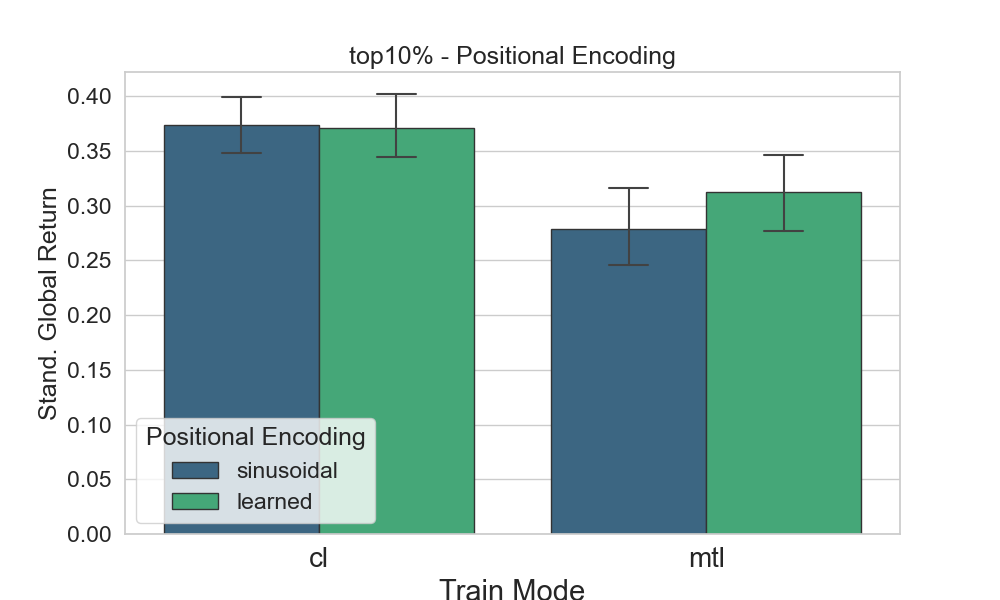} 
        \label{fig:sub2}
    \end{subfigure}
    
    \medskip
    
    \begin{subfigure}[b]{0.45\textwidth}
        \includegraphics[width=\linewidth]{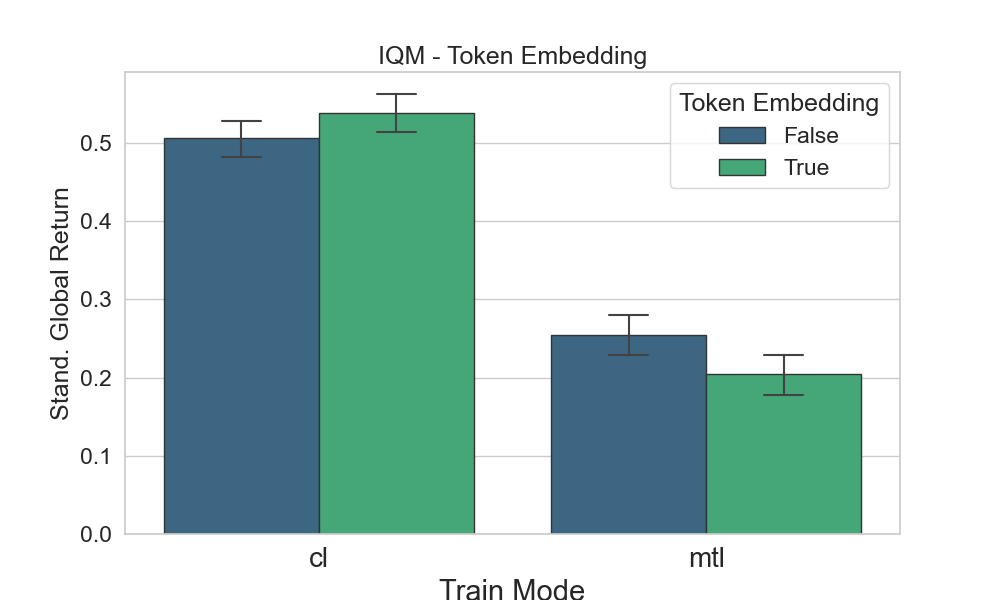} 
        \label{fig:sub1}
    \end{subfigure}
    \hfill
    \begin{subfigure}[b]{0.45\textwidth}
        \includegraphics[width=\linewidth]{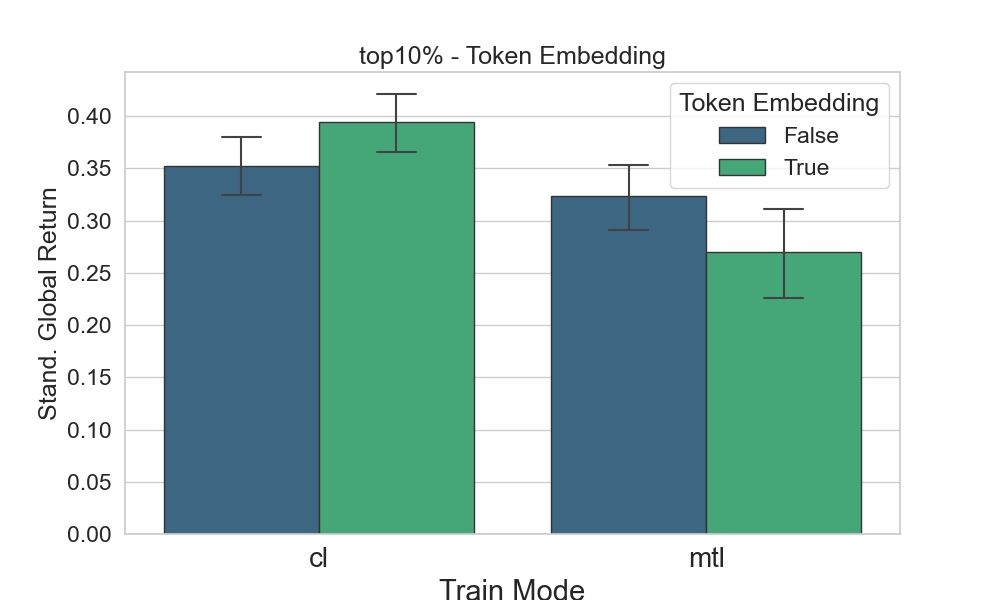} 
        \label{fig:sub2}
    \end{subfigure}
    
    \medskip

\caption{\small{
\textbf{Hyperparameter Sensitivity Analysis for Transformer-based Approaches:} We examine the impact of various hyperparameters on the transformer-based approaches, including the hidden state size, type of positional encoding, learnable token embeddings, and number of attention heads. The analysis focuses on both robustness (IQM on the left) and maximal performance (top 10\% on the right). Consistent with all presented synthetic data results, the outcomes are standardized for each setting. Error bars represent the standard error. Among the hyperparameters, the hidden state size is the only one that significantly affects the results.
}}
\label{fig:tx_hparam_analysis}
\end{figure}

See \figref{fig:tx_hparam_analysis} for an hyperparameter sensitivity analysis of the type of positional encoding, learnable token embedding, the number of attention heads, and the transformer hidden state size.

\subsection{Complete results contrasting CL and MTL in the synthetic data experiments}
\label{app:synthetic_cl_mtl_complete}

\begin{figure}[h!]
\centering
    \begin{subfigure}[b]{0.49\textwidth}
        \includegraphics[width=\linewidth]{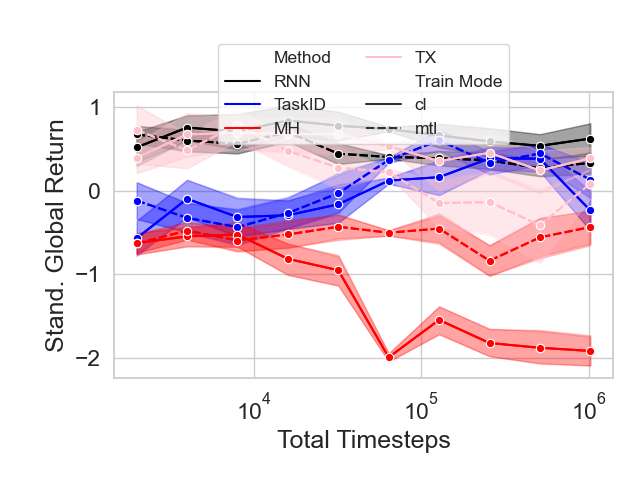} 
        \label{fig:sub1}
    \end{subfigure}
    \hfill
    \begin{subfigure}[b]{0.49\textwidth}
        \includegraphics[width=\linewidth]{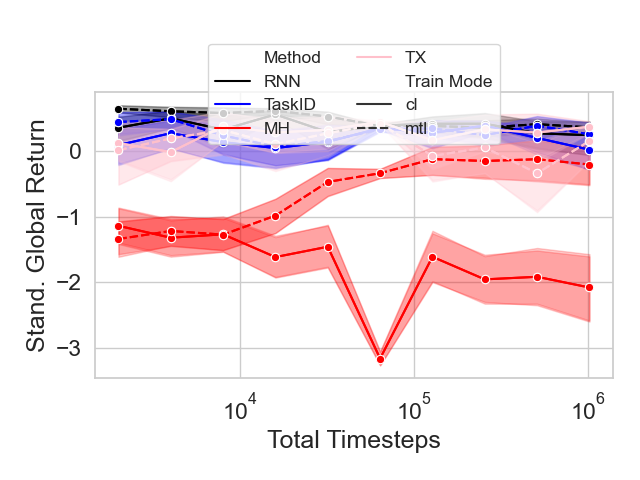} 
        \label{fig:sub2}
    \end{subfigure}
    
    \medskip
    
    \begin{subfigure}[b]{0.49\textwidth}
        \includegraphics[width=\linewidth]{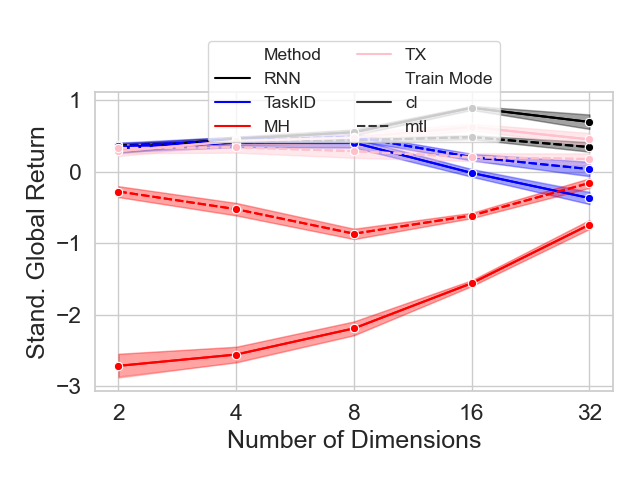} 
        \label{fig:sub1}
    \end{subfigure}
    \hfill
    \begin{subfigure}[b]{0.49\textwidth}
        \includegraphics[width=\linewidth]{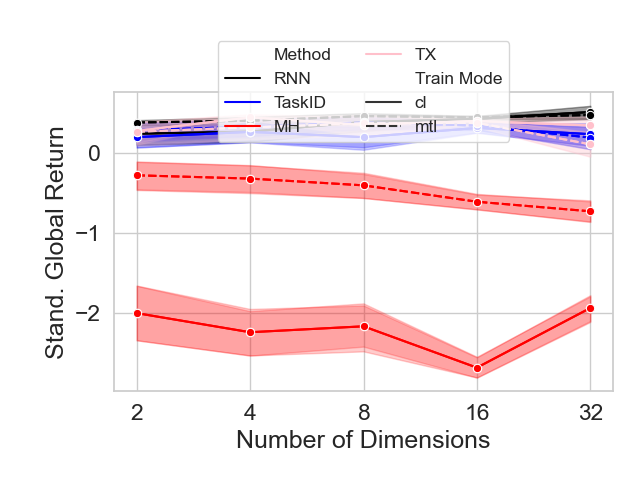} 
        \label{fig:sub2}
    \end{subfigure}
    
    \medskip
    
    \begin{subfigure}[b]{0.49\textwidth}
        \includegraphics[width=\linewidth]{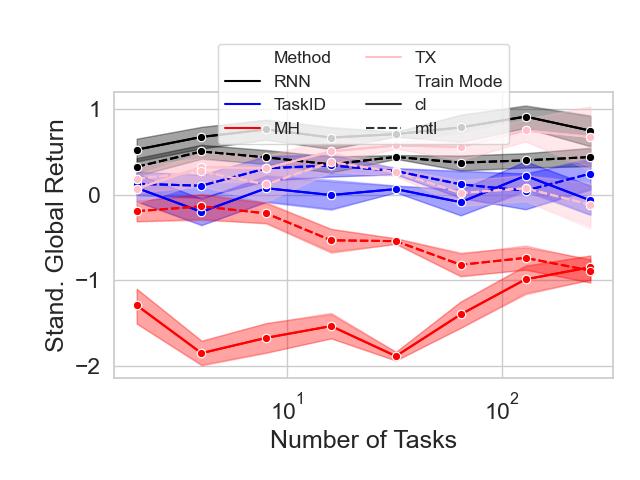} 
        \label{fig:sub1}
    \end{subfigure}
    \hfill
    \begin{subfigure}[b]{0.49\textwidth}
        \includegraphics[width=\linewidth]{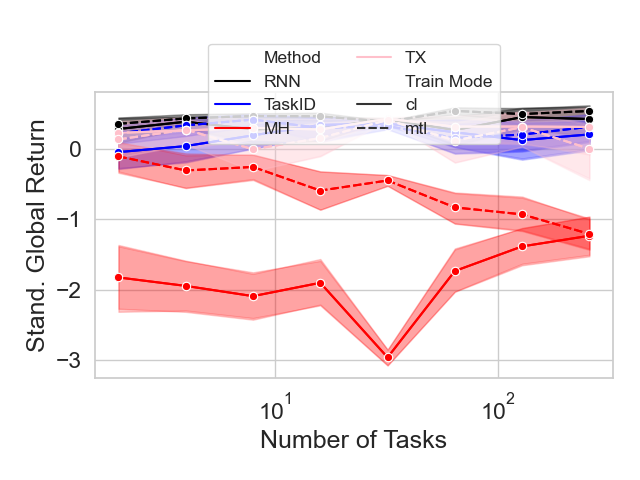} 
        \label{fig:sub2}
    \end{subfigure}
    
\caption{\small{
    \textbf{Complete results for the CL vs MTL synthetic experiments (80k runs).} On the left are IQM plots whereas on the right top 10\% ones.
}}
\label{fig:synthetic_cl_mtl_complete}
\end{figure}

See \figref{fig:synthetic_cl_mtl_complete} for the complete results contrasting CL and MTL in the synthetic data experiments. 

\subsection{More correlation matrices in the synthetic experiments}
\label{app:synthetic_corr_matrices}

\begin{figure}[h!]
\centering

    \begin{subfigure}[b]{0.32\textwidth}
        \includegraphics[width=\linewidth]{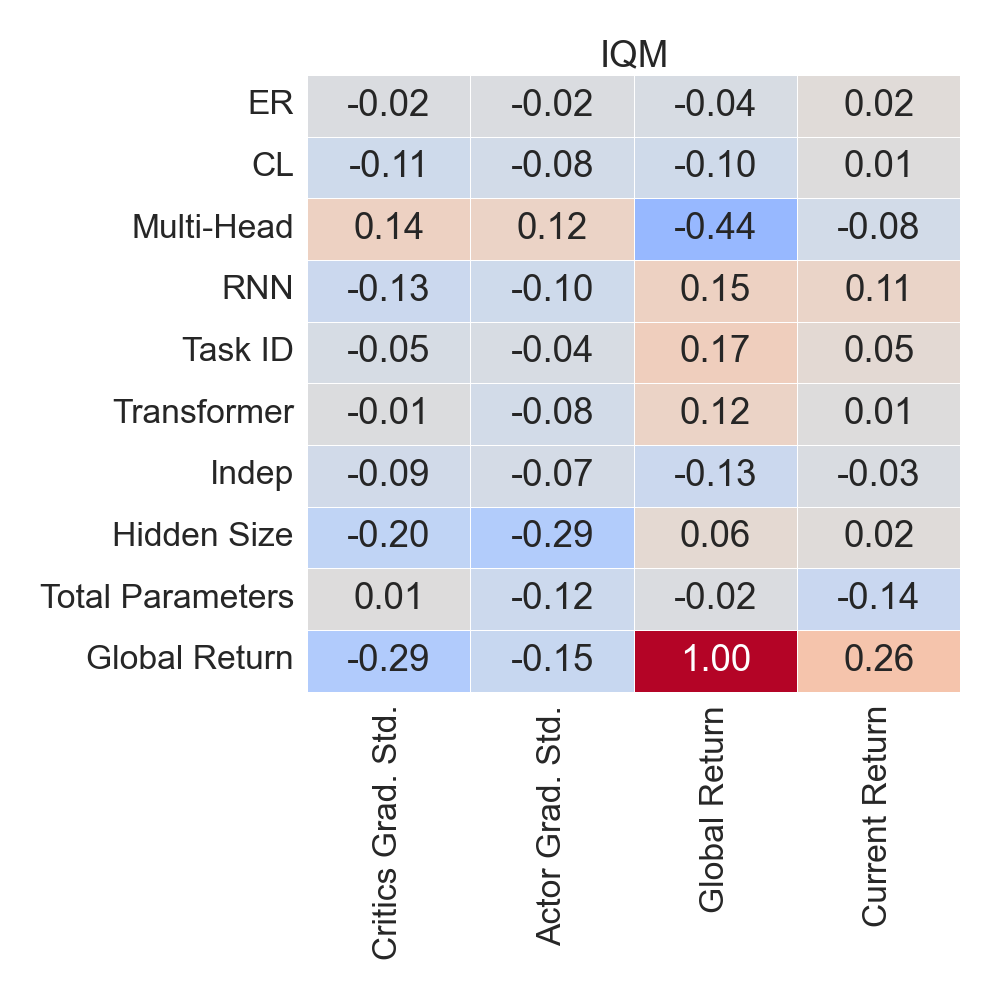} 
        \caption{All runs}
        \label{fig:sub1}
    \end{subfigure}
    \hfill
    \begin{subfigure}[b]{0.32\textwidth}
        \includegraphics[width=\linewidth]{figures/quad_opt/CorrMatrix_IQM-standard_Hardest1M_CL}
        \caption{Challenging CL scenario}
        \label{fig:sub2}
    \end{subfigure}
    \hfill
    \begin{subfigure}[b]{0.32\textwidth}
        \includegraphics[width=\linewidth]{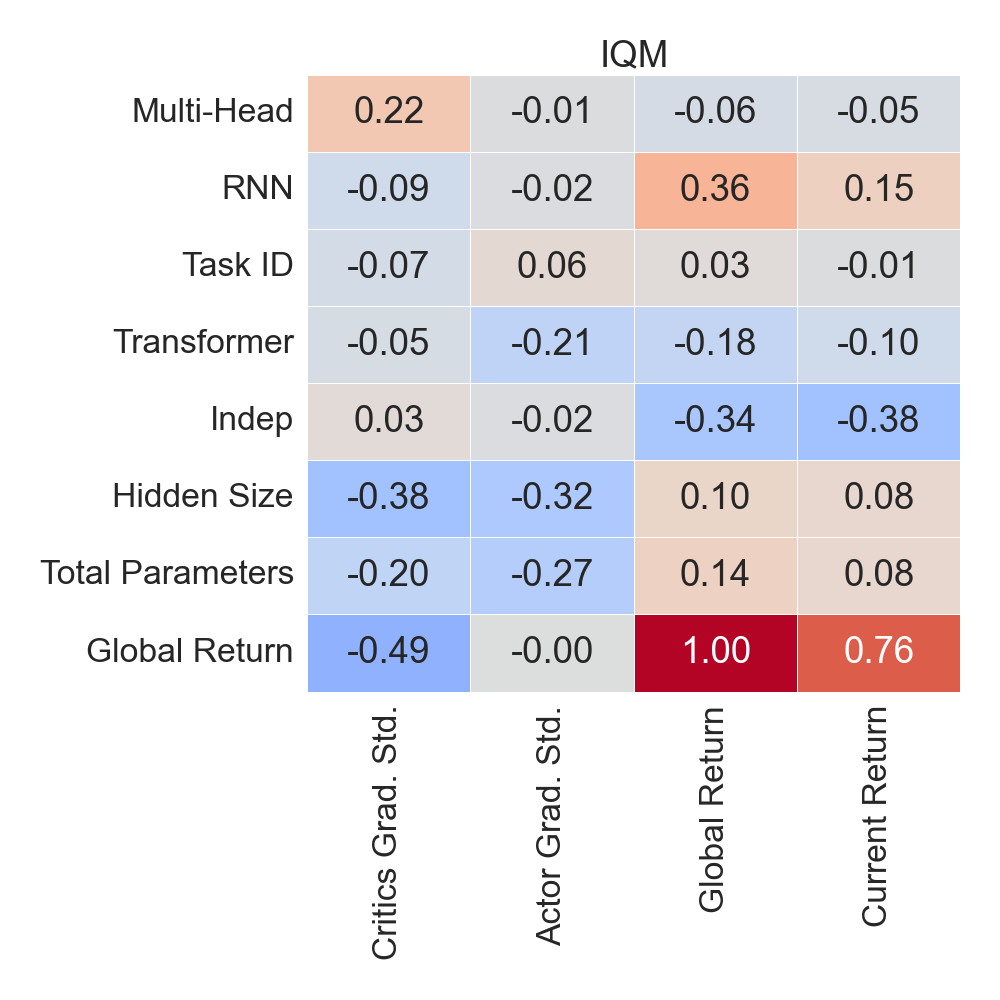}   
        \caption{Challenging MTL scenario}
        \label{fig:sub3}
    \end{subfigure}
    
    \medskip
    
    \begin{subfigure}[b]{0.32\textwidth}
        \includegraphics[width=\linewidth]{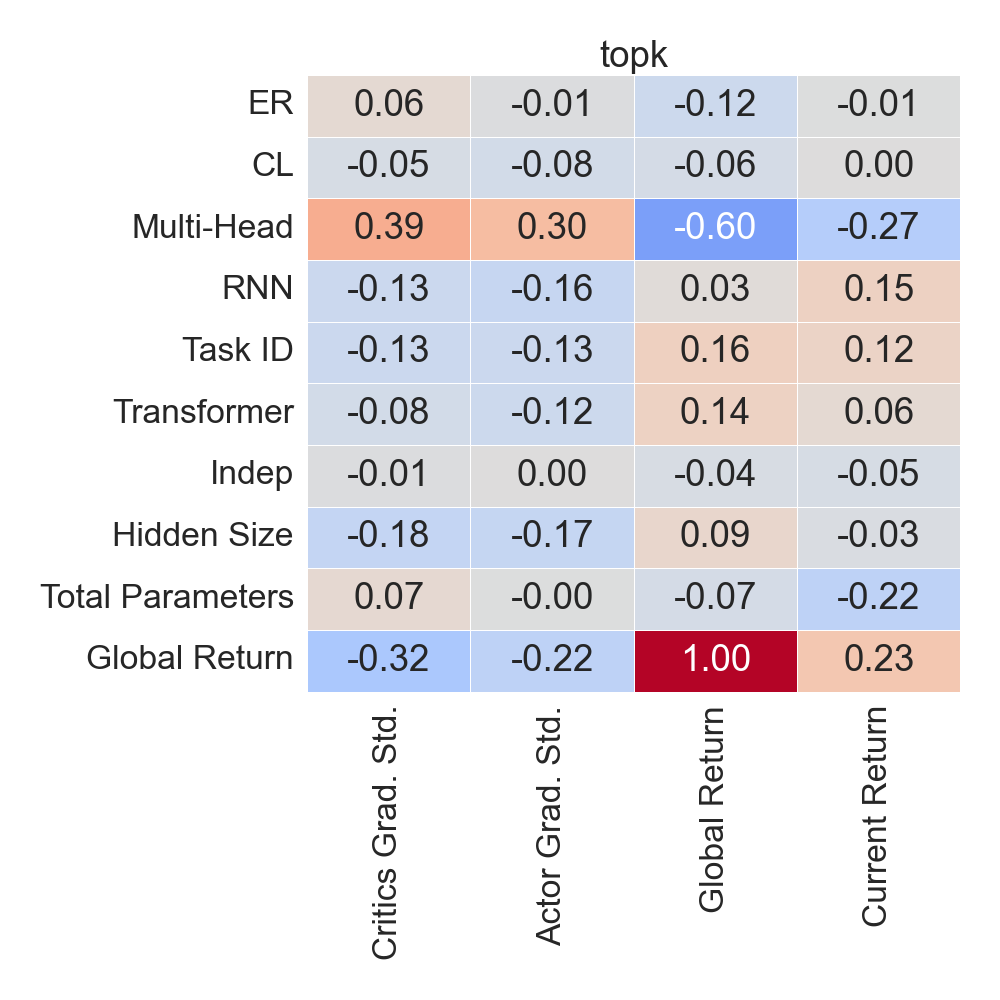} 
        \caption{All runs}
        \label{fig:sub1}
    \end{subfigure}
    \hfill
    \begin{subfigure}[b]{0.32\textwidth}
        \includegraphics[width=\linewidth]{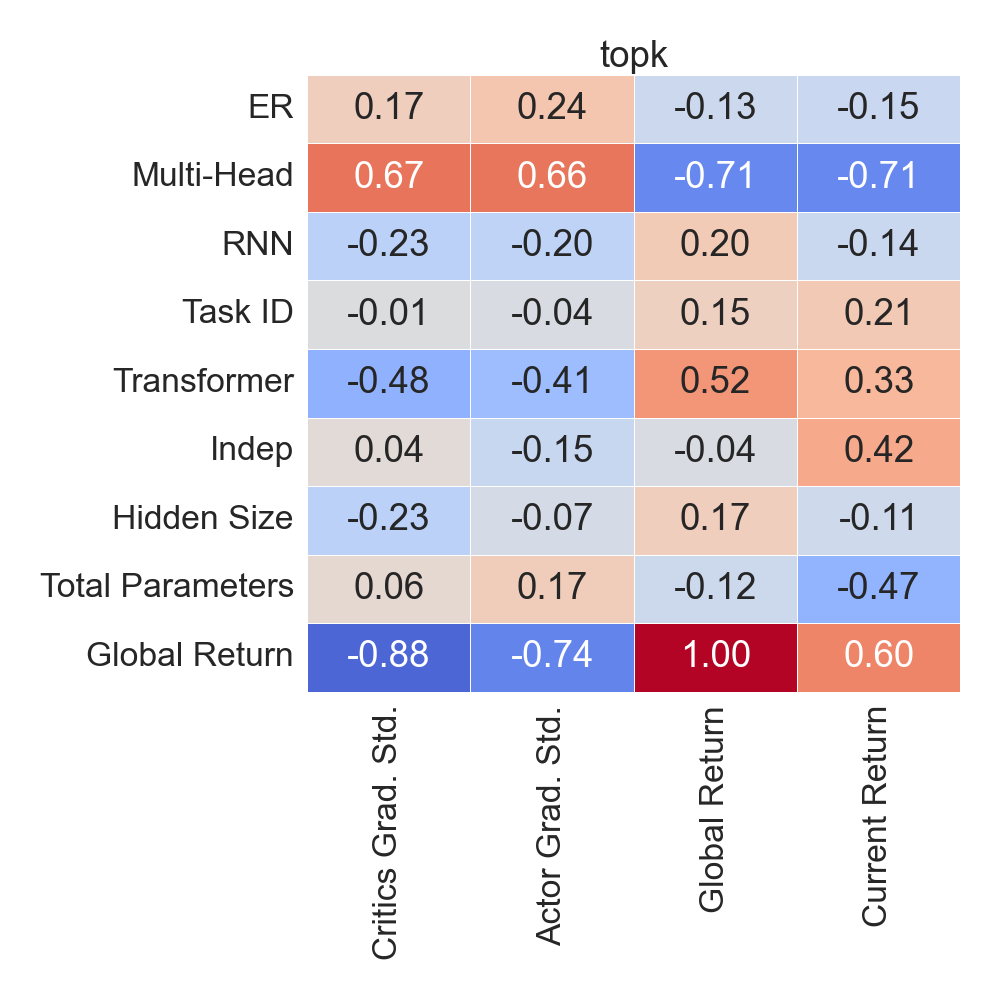}
        \caption{Challenging CL scenario}
        \label{fig:sub2}
    \end{subfigure}
    \hfill
    \begin{subfigure}[b]{0.32\textwidth}
        \includegraphics[width=\linewidth]{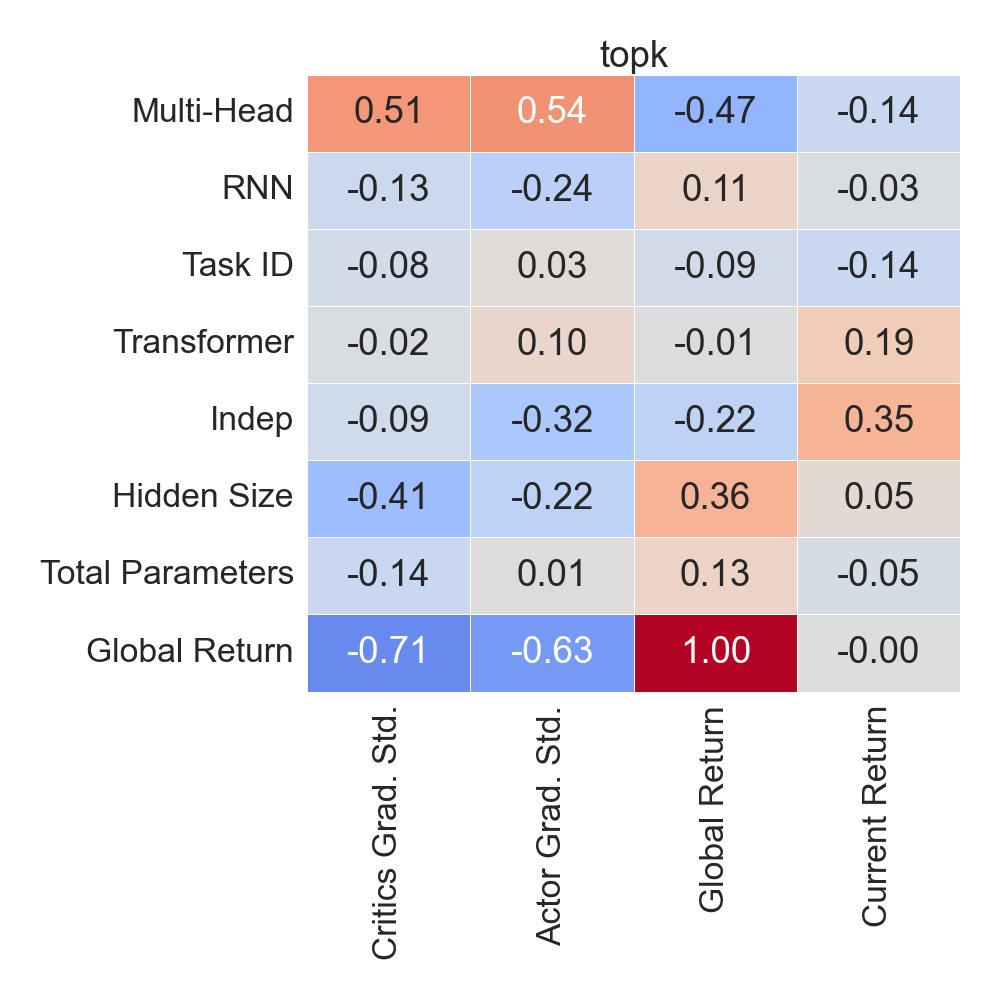}   
        \caption{Challenging MTL scenario}
        \label{fig:sub3}
    \end{subfigure}
\caption{\small{
    \textbf{Spearman correlation matrix for the synthetic experiments (80k runs).} On top are the IQM plots, whereas top 10\% ones are at the bottom.
}}
\label{fig:complete_corr_matrices}
\end{figure}

See \figref{fig:complete_corr_matrices} for more correlation matrices in the synthetic experiment.

  %%%%%%%%%%%%%%%%%%%%%%%%%%%%%%%%%%%%%%%%%%%%%%%%%%%
  \section{Validating our SAC implementation on MT10}
  %%%%%%%%%%%%%%%%%%%%%%%%%%%%%%%%%%%%%%%%%%%%%%%%%%%
  \label{app:MT10}
  
\begin{figure}[h!]
    \centering
          \includegraphics[width=0.6\linewidth]{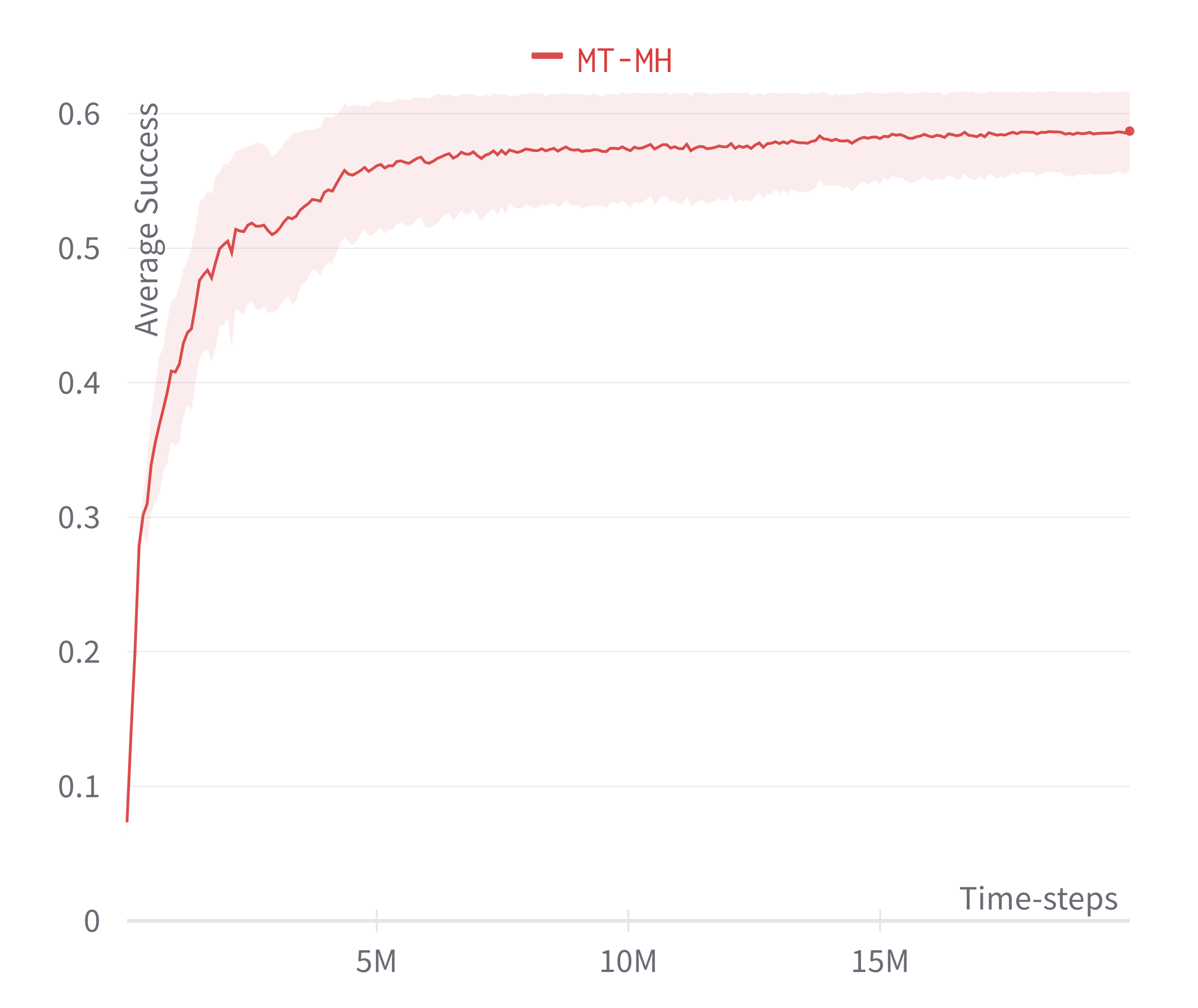} 
      \caption{\textbf{MT10 experiment}. 
      We repeat the popular MT10 benchmark with our MT-MH implementation.
      After 20M time-steps, the algorithm reaches a success rate of 58\%.
      This is in line with Meta-World reported results. 
      In Figure 15 of their Appendix, their MT-SAC is trained for 200M time-steps the first 20M time-steps are aligned with our curve. 
      We further note that our CW10 result might seem weak vis à vis the reported ones in \cite{wolczyk2021continual}. 
      This is explained by \cite{wolczyk2021continual} using Meta-World-v1 instead of the more recent Meta-World-v2. 
      }
      \label{fig:mt10}
  \end{figure}

  In \Figref{fig:mt10} we validate our SAC implementation on Meta-World v2's MT10.
  
%   \section{Why Multi-head achieves poor performance in the synthetic benchmark}
%   \label{app:multi_head_bad}

  %%%%%%%%%%%%%%%%%%%%%%%%%%%%%%%%%%%%%%%%
  \section{Observation in Meta-World \#1: Larger networks do not improve performance in Meta-World}
  %%%%%%%%%%%%%%%%%%%%%%%%%%%%%%%%%%%%%%%%
  \label{app:larger_nets}
  
\begin{figure}[h!]
    \centering
          \includegraphics[width=0.7\linewidth]{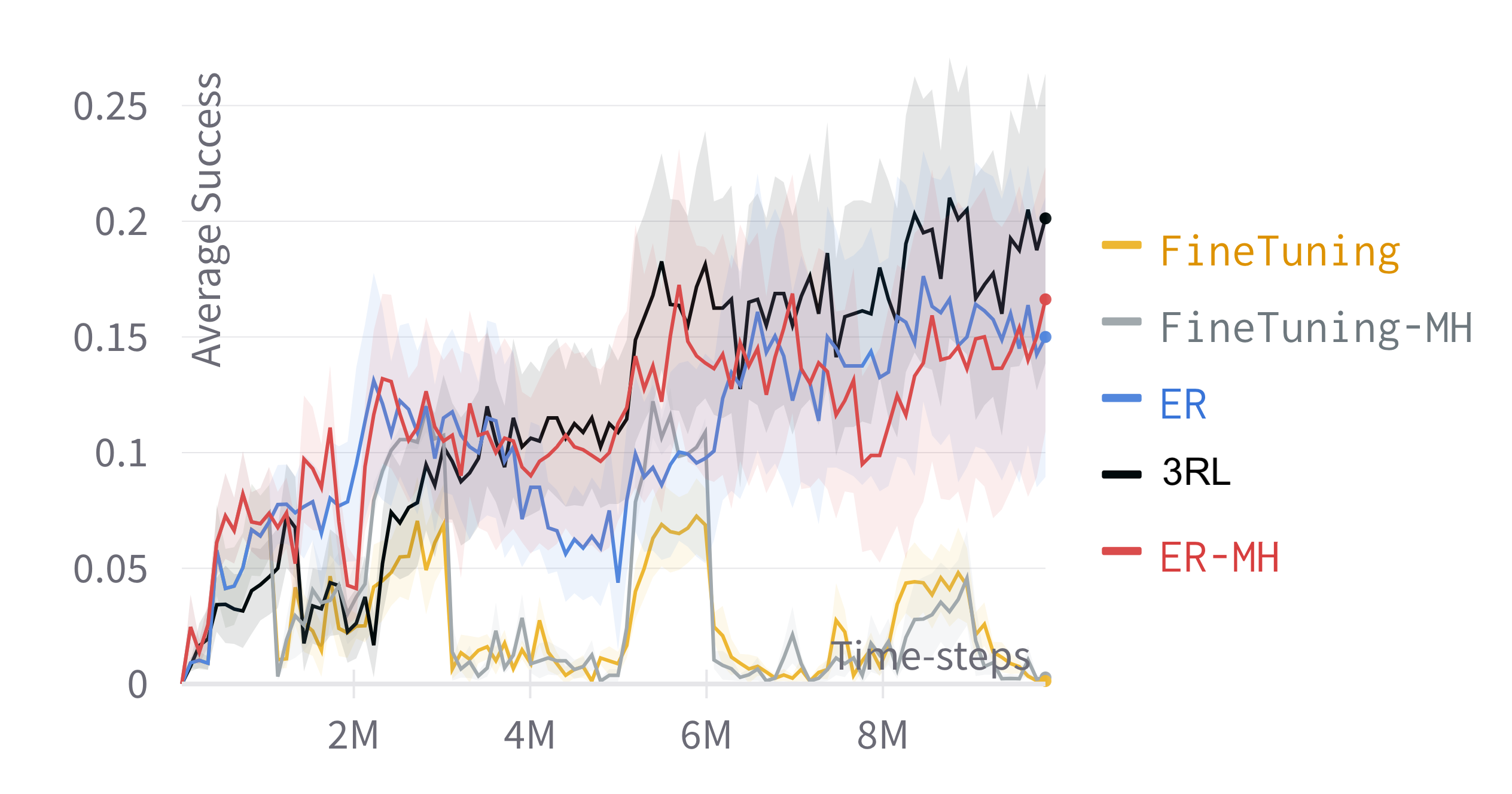} 
      \caption{\textbf{CW10 experiment with larger neural networks.}. 
      We repeated the CW10 experiment, this time with larger neural networks.
      Specifically, we added a third layer to the actor and critics.
      Its size is the same as the previous two, i.e. 400.
      The extra parameters have hindered the performance of all baselines.
          Note that it is not impossible that more well-suited hyperparameters could increase the performance of the bigger networks. 
      }
      \label{fig:cw10_bigger}
  \end{figure}

  In \figref{fig:cw10_bigger} we report that increasing the neural net capacity does not increase performance.

  %%%%%%%%%%%%%%%%%%%%%%%%%%%%%%%%%%%%%%%%%%%
  \section{Observations in Meta-World \#2: RNN doesn't individually improves the single-task performance}
  %%%%%%%%%%%%%%%%%%%%%%%%%%%%%%%%%%%%%%%%%%%
  \label{app:stl}

\begin{figure}[h!]
    \centering
          \includegraphics[width=0.6\linewidth]{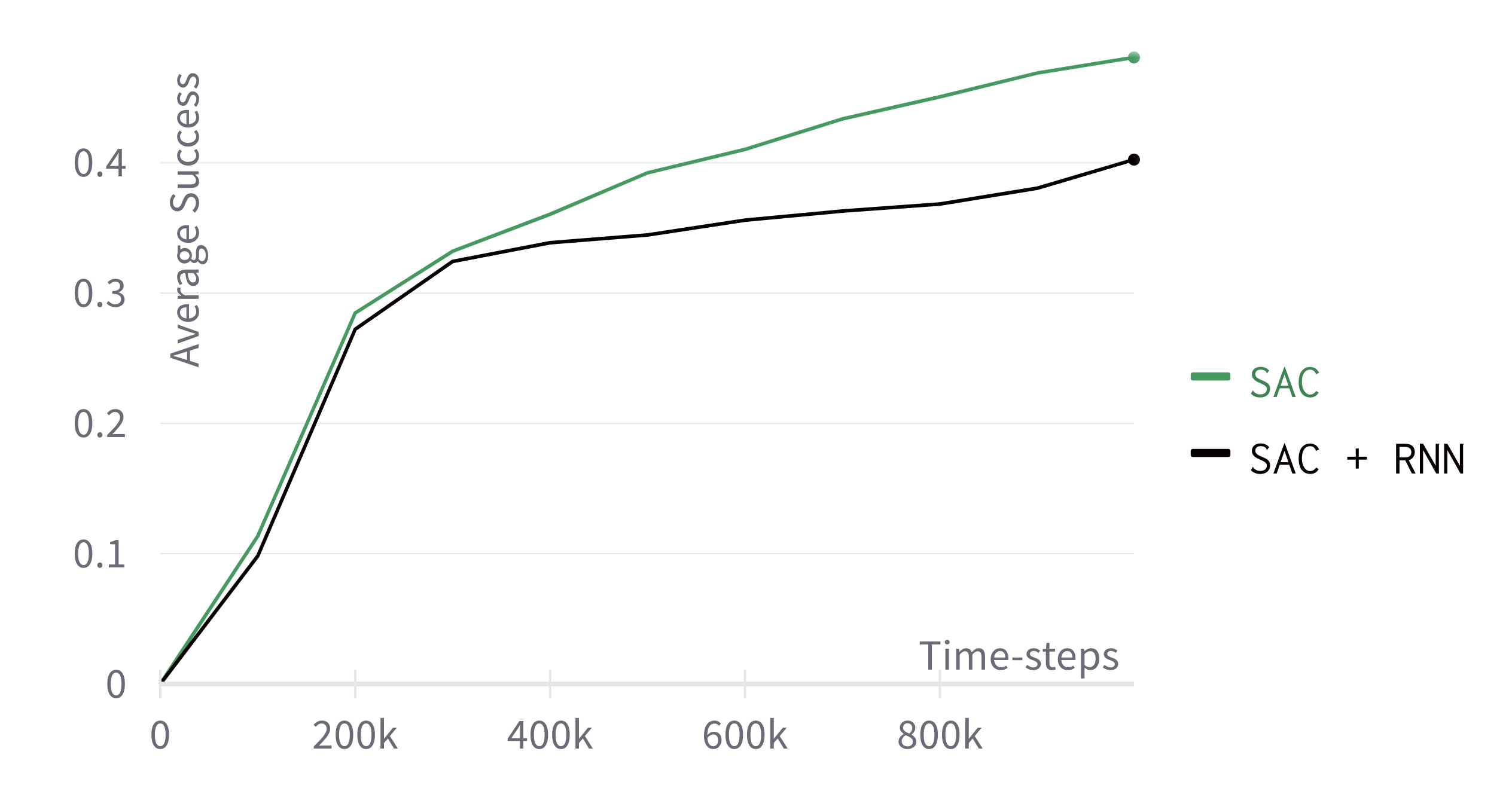} 
      \caption{\textbf{Single-task learning CW10 experiments}.
      Average success on all task trained independently.
      In this regime, the RNN does not help.
      }
      \label{fig:cw10_single}
  \end{figure}

  A simple explanation is that the RNN enhances SAC's ability to learn each task independently, perhaps providing a different inductive bias beneficial to each individual task. 
  To test this hypothesis, we run the \codeword{CW10} benchmark this time with each task learned separately, which we refer to as the Independent baselines.
  Note that this hypothesis is unlikely since the agent observes the complete state which is enough to act optimally (i.e.\ the environments are MDPs not POMDPs). 
  Unsurprisingly, we find the RNN decreases the performance of standard SAC by 8.2\% on average on all tasks. 
  Accordingly, we discard this hypothesis.
  \figref{fig:cw10_single} provides the complete STL results.

  % \begin{figure}[h!]
  %   \centering
  %         \includegraphics[width=0.6\linewidth]{figures/CW10_singles_no_gc.png} 
  %     \caption{\textbf{Single-task learning CW10 experiments without gradient clipping}. 
  %     We enstored gradient clipping in CRL and MTRL to alleviate the deadly triad problem, a problem we did not find in single-task learning (STL). 
  %     For completeness, we reran the STL experiments without gradient clipping.
  %     We found the performance of the RNN to dramatically increase.
  %     Note that this is not the same algorithms used in the CRL and MTRL experiments because of the gradient clipping discrepancy.
  %     }
  %     \label{fig:cw10_single_no_gc}
  % \end{figure}

  %--------------------------------------------------------------------------------------
  \section{Observation in Meta-World \#2: RNN doesn't increases parameter stability}
  \label{app:param_stability}
  
%   \todo[inline]{revise}
  
  The plasticity-stability tradeoff is at the heart of continual learning: plasticity eases the learning of new tasks. 
  Naive learning methods assume stationary data and so are too plastic in non-stationary regimes leading to catastrophic forgetting. To increase stability, multiple methods enforce \citep{mallya2018packnet} or regularize for \citep{kirkpatrick2017overcoming} \emph{parameter stability}, i.e., the tendency of a parameter to stay within its initial value while new knowledge is incorporated. 
  Carefully tuned task-aware methods, e.g. PackNet~\citep{mallya2018packnet} in \cite{ContinualWorld}, have the ability to prevent forgetting.\footnote{The observation that PackNet outperforms \emph{an} MTRL baseline in \cite{ContinualWorld} is different from our stronger observation that a \emph{single} method, namely \algname{}, achieves the same performance in CRL than in MTRL}
  
  Considering the above, we ask: could \algname{} implicitly increase parameter stability?
  To test this hypothesis we measure the average total movement of the each weight throughout an epoch of learning which is defined by all updates in between an episode collection.
  To produce a metric suitable for comparing different runs which could operate in different regimes of parameter updates, we suppose report an entropy-like metric computed as follow $H_e = - (|\theta_{e+1}-\theta_{e}|)^\intercal \log (|\theta_{e+1}-\theta_{e}|)$, where $e$ is the epoch index.
  Note that the proposed metric takes all its sense when used relatively, (i.e. to compare the relative performance in-between methods) and not as an absolute measure of parameter stability 
  Details about this experiment are found in \appref{app:grad_entropy}.

\begin{figure}
    \centering
      \includegraphics[width=0.8\linewidth]{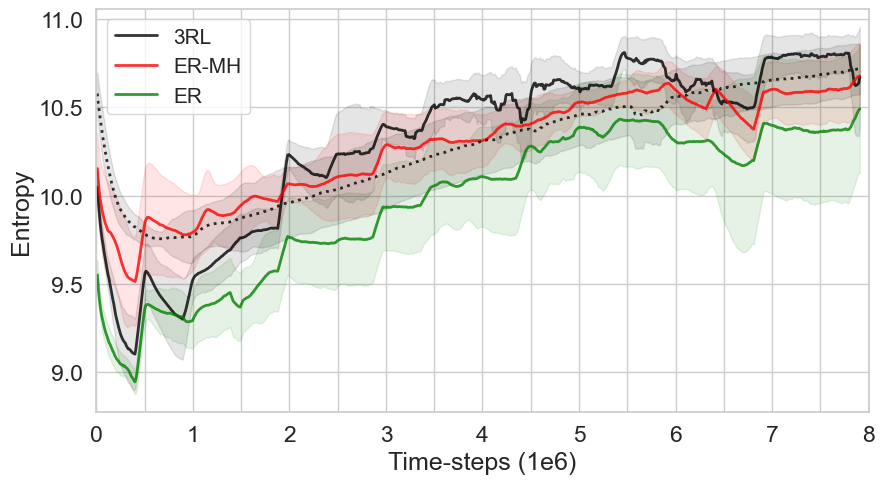} 
    \captionof{figure}{\small{\textbf{3RL does not achieve superior continual learning performance through increased parameter stability}. 
      We show the evolution of the methods' entropy in the parameters updates.
      We include MTL-RNN (dotted line) as a ref.
      We do not observe an increase in parameter stability: on the contrary, all methods, increasingly update more weights as new tasks (or data) come in.
      }}
    \label{fig:grad_entropy}
  \end{figure}

  \Figref{fig:grad_entropy} reports our proxy of total parameter movement for \algname{}, ER, and ER-MH.
  We use these baselines since the gap between ER and its upper bound is the largest and the ER-MH gap is in between ER's and \algname's. 
  We find strong evidence to reject our hypothesis.
  After an initial increase in parameter stability, weight movement increases as training proceeds across all methods and even spikes when a new task is introduced (every 500K steps). MTL-RNN follows the same general pattern as the ER methods.
  Note that, it is not impossible that the \emph{function} represented by the neural networks are stable even though their parameters are not. 
  Nevertheless, it would be surprising that such a behaviour arose only in \algname{} given the high similarity of the stability reported across all methods.

%   \begin{figure}
%   \centering
%   \begin{minipage}{.53\textwidth}
%     \centering
%           \includegraphics[width=\linewidth]{figures/MW/CL_MTL_MW20_500k_90_int.png} 
%     \captionof{figure}{\small{\textbf{3RL reaches its MTRL soft-upper bound}.
%           Continual vs multi-task learning methods in solid lines and dotted lines, respectively.
%           \algname{} methods matches its soft-upper bound MTL analog as well as the other MTRL baselines. 
%           In contrast, other baselines' performance are drastically hindered by the non-stationary task distribution.
%       }}
%     \label{fig:mw20-500k}
%   \end{minipage}%
%   \hspace{8pt}
%   \begin{minipage}{.44\textwidth}
%     \centering
%       \includegraphics[width=\linewidth]{figures/MW/MW20_actor_entropy_90_int.png} 
%     \captionof{figure}{\small{\textbf{3RL does not achieve superior continual learning performance through increased parameter stability}. 
%       We show the evolution of the methods' entropy in the parameters updates.
%       We include MTL-RNN (dotted line) as a ref.
%       We do not observe an increase in parameter stability: on the contrary, all methods, increasingly update more weights as new tasks (or data) come in.
%       }}
%     \label{fig:grad_entropy}
%   \end{minipage}
%   \end{figure}

\section{Observation in Meta-World \#3: RNN correctly places the new tasks in the context of previous ones, enabling forward transfer and improving optimization}
\label{app:recontext}

% \todo[inline]{revise}

As in real robotic use-cases, MW tasks share a set of low-level reward components like grasping, pushing, placing, and reaching, as well as set of object with varying joints, shapes, and connectivity.
As the agent experiences a new task, the RNN could quickly infer how the new data distribution relates with the previous ones and provide to the actor and critics a useful representation.
Assume the following toy example: task one's goal is to grasp a door handle, and task two's to open a door.
The RNN could infer from the state-action-reward trajectory that the second task is composed of two subtasks: the first one as well as a novel pulling one.
Doing so would increase the policy learning's speed, or analogously enable forward transfer. 

Now consider a third task in which the agent has to close a door. 
Again, the first part of the task consists in grasping the door handle.
However, now the agent needs to subsequently push and not pull, as was required in task two. 
In this situation, task interference \cite{yu2020gradient} would occur.
Once more, if the RNN could dynamically infer from the context when pushing or pulling is required,
it could modulate the actor and critics to have different behaviors in each tasks thus reducing the interference. Note that a similar task interference reduction should be achieved by task-aware methods.
E.g., a multi-head component can enable a method to take different actions in similar states depending on the tasks, thus reducing the task interference.

Observing and quantifying that \algname{} learns a representation space in which the new tasks are correctly \emph{decomposed} and placed within the previous ones is challenging.  
Our initial strategy is to look for effects that should arise if this hypothesis was true (so observing the effect would confirm the hypothesis).

\begin{figure}
    \centering
    \begin{minipage}{0.5\linewidth}
        \includegraphics[trim={0.cm 0 0 0}, width=1.025\linewidth]{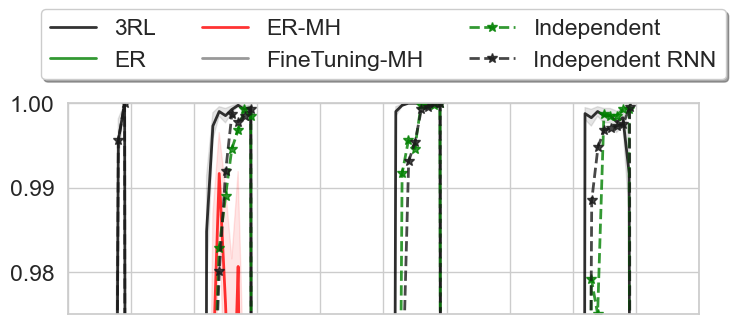}
        \includegraphics[width=\linewidth]{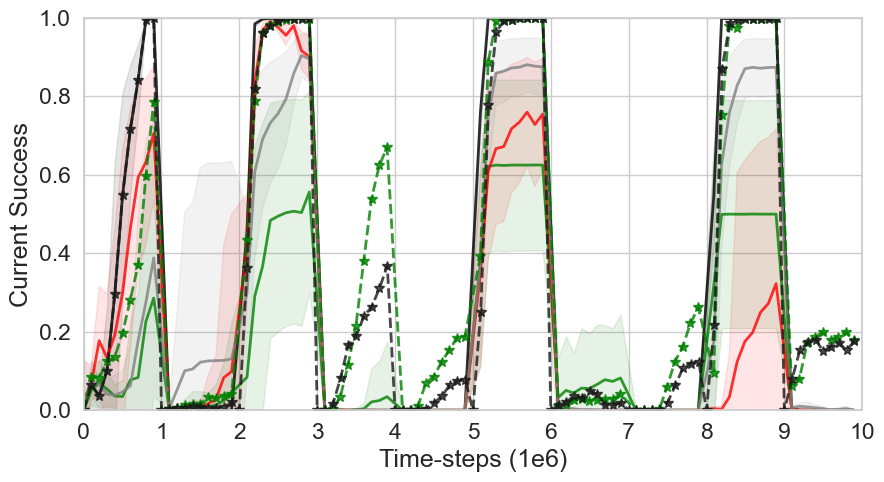}
        \captionof{figure}{\small{\textbf{\algname{} is the fastest continual learner.}
        Current success rate on \texttt{\small\textcolor{blue}{CW10}}. 
        The Independent method, which trains on each task individually, is still the best approach to maximise single-task performance.
        However, on the task that the continual learning methods succeed at, \algname{} is the fastest learner (see the zoom in the top plot). 
        In these cases, its outperformance over Independent and Independent RNN indicates that forward transfer is achieved. 
        }}
        \label{fig:cw10_current}
    \end{minipage}
\end{figure}

First, we take a look at the time required to adapt to new tasks: if \algname{} correctly infers how new tasks relate to previous ones, it might be able to learn faster by re-purposing learned behaviors.
\Figref{fig:cw10_current} depicts the current performance of different methods throughout the learning of \codeword{CW10}.
For reference, we provide the results of training separate models, which we refer to as Independent and Independent RNN. 

The challenges of Meta-World v2 compounded with the ones from learning multiple policies in a shared network, and handling an extra level of non-stationary, i.e. in the task distribution, leaves the continual learners only learning task 0, 2, 5, and 8. 
On those tasks (except the first one in which no forward transfer can be achieved) \algname{} is the fastest continual learner.
Interestingly, \algname{} showcases some forward transfer by learning faster than the Independent methods on those tasks. 
This outperformance is more impressive when we remember that \algname{} is spending most of its compute replaying old tasks. 
We thus find some support for Hypothesis \#3.

Second, simultaneously optimizing for multiple tasks can lead to conflicting gradients or task interference \citep{yu2020gradient}. To test for this effect, we use the average variance of the gradients on the mini-batch throughout the training as a proxy of gradient conflict.
We explain in \ref{app:grad_variance} why we use the gradients' variance to measure  gradient conflict instead of using the angle between the gradients as in \cite{yu2020gradient}.

In \figref{fig:stability} we show the normalized global success metric plotted against the gradient variance. 
In line with our intuition, we do find that the RNN increases gradient agreement over baselines.
As expected, adding a multi-head scheme can also help, to a lesser extent. 
We find a significant negative correlation of -0.75 between performance and gradient conflict.
\appref{app:grad_conflict} reports the evolution of gradient conflict through time in the actor and critic networks.

\Figref{fig:stability} also reports training stability, as measured by the standard deviation of Q-values throughout training (not to be confused with the parameter stability, at the center of Hypothesis \#2, which measures how much the parameters move around during training).
We find \algname{} enjoys more stable training as well as a the significant negative correlation of -0.81 between performance and training stability. 
Note that 
The plausibility of Hypothesis \#3 is thus further increased.

We wrap up the hypothesis with some qualitative support for it.
\Figref{fig:fig1} showcases the RNN representations as training unfolds.
If the RNN was merely performing task inference, we would observe the trajectories getting further from each other, not intersecting, and collapsing once the task is inferred. 
Contrarily, the different task trajectories constantly evolve and seem to intersect in particular ways. 
Although only qualitative, this observation supports the current hypothesis.

  %%%%%%%%%%%%%%%%%%%%%%%%%%%%%%%%%%%%%%%%
  \section{Gradient Entropy Experiment}
  %%%%%%%%%%%%%%%%%%%%%%%%%%%%%%%%%%%%%%%%
  \label{app:grad_entropy}
  
  To assess parameter stability, we look at the entropy of the parameters for each epoch.
  Because the episodes are of size 500, the epoch corresponds to 500 updates.
  To remove the effect of ADAM~\citep{kingma2017adam}, our optimizer, we approximate the parameters' movement by summing up their absolute gradients throughout the epoch.
  To approximate the sparsity of the updates, we report the entropy of the absolute gradient sum. 
  For example, a maximum entropy would indicate all parameters are moving equally.
  If the entropy drops, it means the algorithm is applying sparser updates to the model, similarly to PackNet.
  
  %%%%%%%%%%%%%%%%%%%%%%%%%%%%%%%%%%%%%%%%
  \section{Gradient variance as a proxy for gradient conflict}
  %%%%%%%%%%%%%%%%%%%%%%%%%%%%%%%%%%%%%%%%
  \label{app:grad_variance}
  
  \cite{yu2020gradient} instead measure the conflict between two tasks via the angle between their gradients, proposing that the tasks conflict if this angle is obtuse. 
  However we argue that it also critical to consider the magnitude of the gradients. 
  
  Consider an example of two tasks involving 1D optimization with two possible cases.  
  In case 1, suppose that the gradient of the first task is 0.01 and the gradient of the second is -0.01. 
  In case 2, suppose that the gradients are now 0.01 and 0.5, respectively. Measuring conflict via the angle would lead us to think that the tasks are in conflict in case 1 and are not in case 2. 
  However, their agreement is actually much higher in case 1: both tasks agree that they should not move too far from the current parameter. 
  In case 2, although the two tasks agree on the direction of the parameter update step, they have loss landscapes with great differences in  curvature. 
  The update step will be too small or too large for one of the  tasks. We thus find variance to be a better measure of gradient conflict.

  %%%%%%%%%%%%%%%%%%%%%%%%%%%%%%%%%%%%%%%%
  \section{Gradient conflict through time}
  %%%%%%%%%%%%%%%%%%%%%%%%%%%%%%%%%%%%%%%%
  \label{app:grad_conflict}
  
\begin{figure}[h]
    \centering
           \includegraphics[width=0.49\linewidth]{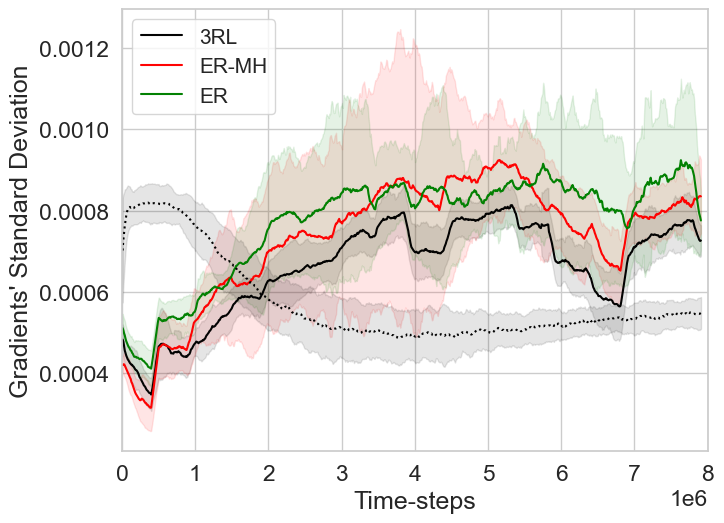}
           \includegraphics[width=0.49\linewidth]{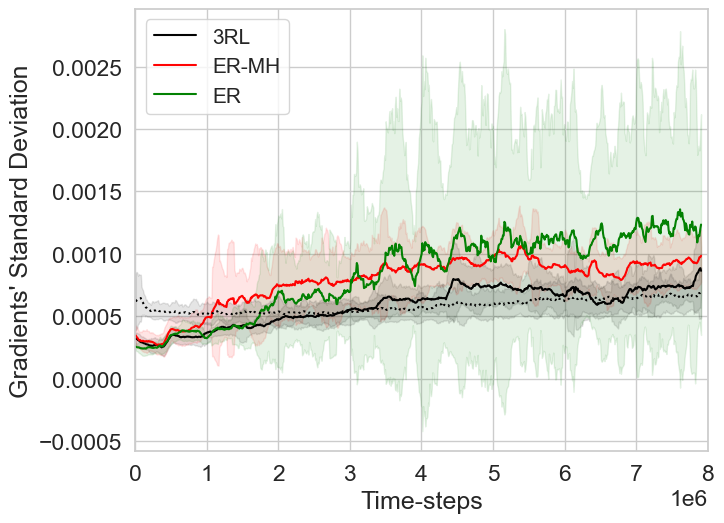} 
      \caption{\textbf{Gradient variance analysis on CW20}. 
      Comparison of the normalized standard deviation of the gradients for the actor (left) and critics (right) in for different CRL methods. 
      For reference, we included MTL-RNN as the dotted line.
      The gradient alignment's rank is perfectly correlated with the performance rank.
      }
      \label{fig:grad_conflict}
  \end{figure}

  See \figref{fig:grad_conflict}.
  
  %%%%%%%%%%%%%%%%%%%%%%%%%%%%%%%%%%%%%%%%
  \section{Task-aware meets task-agnostic}
  %%%%%%%%%%%%%%%%%%%%%%%%%%%%%%%%%%%%%%%%
  \label{app:er-rnn-mh}
  
\begin{figure}[h!]
    \centering
          \includegraphics[width=0.6\linewidth]{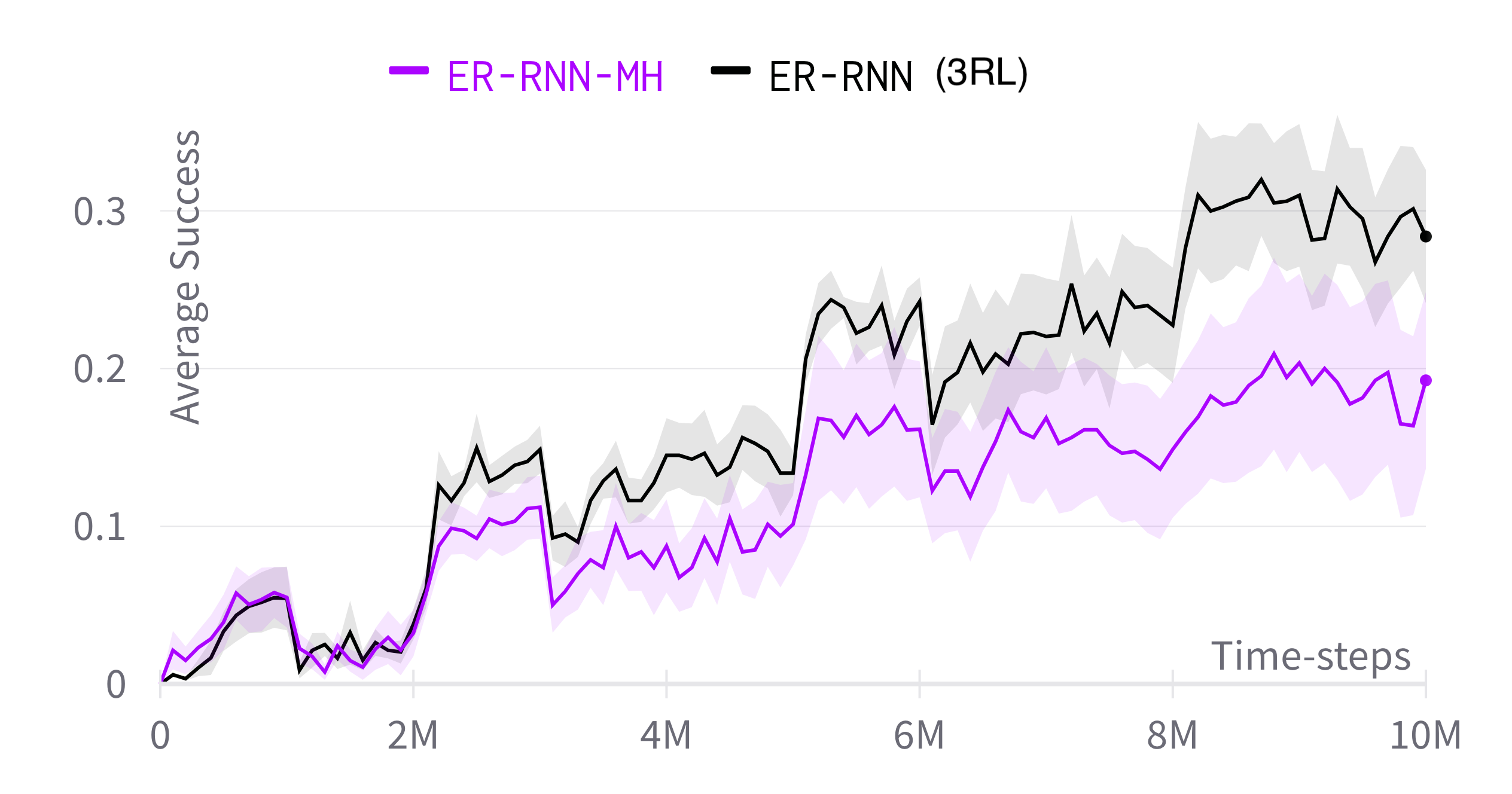} 
      \caption{\textbf{CW10 experiment combining the RNN and MH}. 
      Combining the best task-agnostic (\algname{}) and task-aware (ER-MH) CRL methods did not prove useful.
      Note that this experiment was ran before we enstored gradient clipping, which explains why the performance is lower than previously reported. 
      }
      \label{fig:cw10_er-rnn-mh}
  \end{figure}

  In \figref{fig:cw10_er-rnn-mh} we show that combining the RNN with MH is not a good proposition in \codeword{CW10}.
  
  %%%%%%%%%%%%%%%%%%%%
  \section{Limitations}
  %%%%%%%%%%%%%%%%%%%%
  \label{sec:limits}

  In this section we surface some limitations in our work.
  First of all, given the highly-demanding computational nature of the studied benchmarks, we could not run extensive hyperparameter searches as well as all the desired ablations.
  We have mostly relied on the hyperparameters prescribed by Meta-World, with the exception of the introduction of gradient clipping (\appref{app:exp_details}) which we found detrimental for the continual and multi-task learners to perform adequately. 
  Importantly, we have not tested multiple context length for \algname{}, an important hyperparameter for model-free recurrent RL~\citep{Tianwei2021recurrent}. 
  We can thus hypothesize that our \algname{}'s performances are underestimation.
  
  Second, although Meta-World is know to be challenging, one could argue that its task-inference component is uncomplicated.
  Future work could explore task-agnostic continual RL benchmarks in which the different MDP are less distinctive. 
  
  Third, the hypothesis about the RNN correctly placing the new tasks in the context of previous ones (Hypothesis \#3) is difficult to test.
  We have tested for effects that should arise if the hypothesis is true, but those effect could arise for different reasons.
  Future work could potentially dismantle the reward functions and look for RNN context overlaps across tasks when a particular reward component, e.g., grasping a door knob, is activated.
  
  Lastly, we have only investigated a single benchmark. 
  Introducing more benchmarks would was outside our compute budget.
  Moreover, challenging RL benchmarks have steep user learning curve, especially in continual RL.
  We leave for future work the study of \algname{} in a wide suite of benchmarks.

\end{document}